\documentclass[10pt,twocolumn,letterpaper]{article}
\newcommand{\OtterArchiveRoot}{./}
\usepackage[OT1]{fontenc}
\PassOptionsToPackage{table}{xcolor}
\usepackage[pagenumbers]{cvpr}
\usepackage{tabularx,array,multirow,pifont,algorithm}
\usepackage{amsthm,mathrsfs,textcomp,algorithmicx,algpseudocode,listings}
\definecolor{cvprblue}{rgb}{0.21,0.49,0.74}
\usepackage[breaklinks,colorlinks,allcolors=cvprblue,hyperfootnotes=false]{hyperref}
\RequirePackage{iftex}
\ifPDFTeX\else
  \newcount\pdfcompresslevel
  \newcount\pdfoptionpdfminorversion
  \newcount\pdfgentounicode
  \providecommand{\pdfcatalog}[1]{}
  \providecommand{\pdfglyphtounicode}[2]{}
\fi

\usepackage{xcolor}
\definecolor{cmlabPrefix}{HTML}{0D207F}  % cmlab dark navy
\definecolor{cmlabNumber}{HTML}{0D207F}  % cmlab dark navy (unified single color)
\definecolor{cmlabCite}{HTML}{0D207F}    % cite also dark navy
\definecolor{cmlabURL}{HTML}{0D207F}     % URL also dark navy

\usepackage{hyperref}
\hypersetup{
  colorlinks=true,
  linkcolor=cmlabNumber,    % Number color for \ref (cref is redefined below)
  citecolor=cmlabCite,      % \cite (references)
  urlcolor=cmlabURL,        % \url, \href (external URLs)
  filecolor=cmlabURL,
  pdfborder={0 0 0},
}

\usepackage[capitalize,noabbrev]{cleveref}

\newcommand{\cmlabRef}[4]{%
  {\color{cmlabPrefix}#1}~#3{\color{cmlabNumber}#2}#4%
}
\newcommand{\cmlabRefParen}[4]{% For equations: wrap number in (...)
  {\color{cmlabPrefix}#1}~#3{\color{cmlabNumber}(#2)}#4%
}

\crefformat{figure}     {\cmlabRef{Fig.}{#1}{#2}{#3}}
\Crefformat{figure}     {\cmlabRef{Figure}{#1}{#2}{#3}}
\crefrangeformat{figure}%
  {{\color{cmlabPrefix}Figs.}~#3{\color{cmlabNumber}#1}#4--#5{\color{cmlabNumber}#2}#6}
\crefmultiformat{figure}%
  {\cmlabRef{Figs.}{#1}{#2}{#3}}%
  {, #2{\color{cmlabNumber}#1}#3}%
  {, #2{\color{cmlabNumber}#1}#3}%
  {, and #2{\color{cmlabNumber}#1}#3}

\crefformat{table}      {\cmlabRef{Tab.}{#1}{#2}{#3}}
\Crefformat{table}      {\cmlabRef{Table}{#1}{#2}{#3}}
\crefrangeformat{table}%
  {{\color{cmlabPrefix}Tabs.}~#3{\color{cmlabNumber}#1}#4--#5{\color{cmlabNumber}#2}#6}
\crefmultiformat{table}%
  {\cmlabRef{Tabs.}{#1}{#2}{#3}}%
  {, #2{\color{cmlabNumber}#1}#3}%
  {, #2{\color{cmlabNumber}#1}#3}%
  {, and #2{\color{cmlabNumber}#1}#3}

\crefformat{section}    {\cmlabRef{Sec.}{#1}{#2}{#3}}
\Crefformat{section}    {\cmlabRef{Section}{#1}{#2}{#3}}
\crefformat{subsection} {\cmlabRef{Sec.}{#1}{#2}{#3}}
\Crefformat{subsection} {\cmlabRef{Section}{#1}{#2}{#3}}
\crefformat{subsubsection}{\cmlabRef{Sec.}{#1}{#2}{#3}}
\Crefformat{subsubsection}{\cmlabRef{Section}{#1}{#2}{#3}}
\crefrangeformat{section}%
  {{\color{cmlabPrefix}Secs.}~#3{\color{cmlabNumber}#1}#4--#5{\color{cmlabNumber}#2}#6}
\crefmultiformat{section}%
  {\cmlabRef{Secs.}{#1}{#2}{#3}}%
  {, #2{\color{cmlabNumber}#1}#3}%
  {, #2{\color{cmlabNumber}#1}#3}%
  {, and #2{\color{cmlabNumber}#1}#3}

\crefformat{equation}   {\cmlabRefParen{Eq.}{#1}{#2}{#3}}
\Crefformat{equation}   {\cmlabRefParen{Equation}{#1}{#2}{#3}}
\crefrangeformat{equation}%
  {{\color{cmlabPrefix}Eqs.}~#3{\color{cmlabNumber}(#1)}#4--#5{\color{cmlabNumber}(#2)}#6}
\crefmultiformat{equation}%
  {\cmlabRefParen{Eqs.}{#1}{#2}{#3}}%
  {, #2{\color{cmlabNumber}(#1)}#3}%
  {, #2{\color{cmlabNumber}(#1)}#3}%
  {, and #2{\color{cmlabNumber}(#1)}#3}

\crefname{algorithm}{Algo.}{Algos.}
\Crefname{algorithm}{Algorithm}{Algorithms}
\crefformat{algorithm}  {\cmlabRef{Algo.}{#1}{#2}{#3}}
\Crefformat{algorithm}  {\cmlabRef{Algorithm}{#1}{#2}{#3}}
\crefrangeformat{algorithm}%
  {{\color{cmlabPrefix}Algos.}~#3{\color{cmlabNumber}#1}#4--#5{\color{cmlabNumber}#2}#6}
\crefmultiformat{algorithm}%
  {\cmlabRef{Algos.}{#1}{#2}{#3}}%
  {, #2{\color{cmlabNumber}#1}#3}%
  {, #2{\color{cmlabNumber}#1}#3}%
  {, and #2{\color{cmlabNumber}#1}#3}

\makeatletter
\providecommand{\input@path}{}
\edef\input@path{{\OtterArchiveRoot}\input@path}
\makeatother
\usepackage[
  enable,
  applybold,
  labname={CMLab},
  titleicon={\OtterArchiveRoot 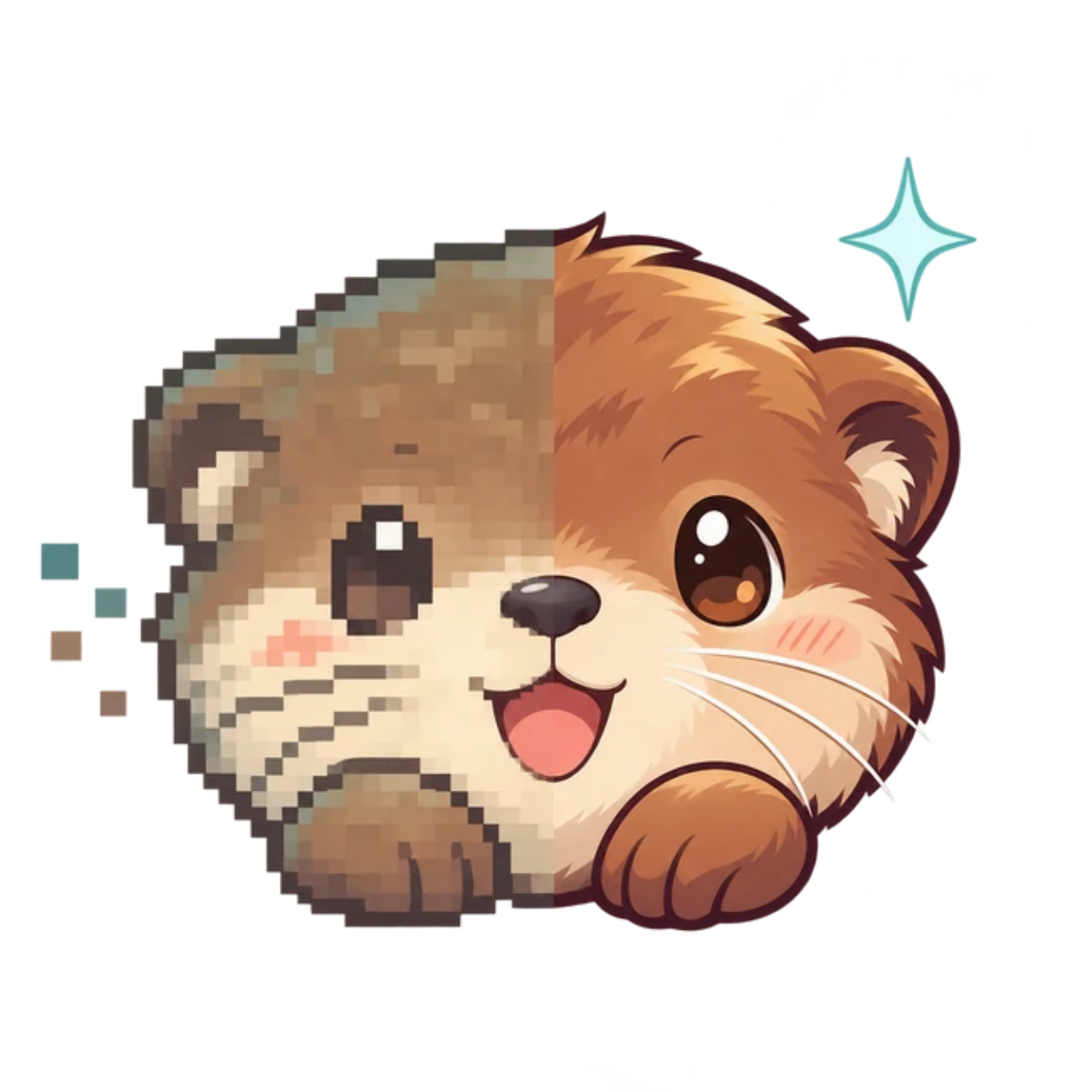},
  logowidth={21mm},
  projectpage={https://cmlab-korea.github.io/OTTER/}
]{cmlab_1}
\newcommand{\OTTERHeadingPatchError}[1]{%
  \PackageError{otter-arxiv}{Could not apply Ubuntu Bold to #1}{Check the CVPR style.}}
\patchcmd{\cvprsection}{\large\bf}{\large\ubuntubold}{}{\OTTERHeadingPatchError{section}}
\patchcmd{\cvprsubsection}{\elvbf}{\fontsize{11}{13}\selectfont\ubuntubold}{}{\OTTERHeadingPatchError{subsection}}
\patchcmd{\cvprsubsubsection}{\tenbf}{\fontsize{10}{12}\selectfont\ubuntubold}{}{\OTTERHeadingPatchError{subsubsection}}
\makeatletter
\newcommand{\OtterArchiveCover}{\cmlab@renderbox}
\makeatother
\catcode`’=\active
\protected\def’{\textquoteright}
\hypersetup{bookmarksnumbered=true,pdfdisplaydoctitle=true}

\newcommand{\PaperTitleLines}[1]{OTTER: One-Step Teacher-Free Image Restoration#1via Nonparametric Score Distillation}
\newcommand{\PaperTitle}{\PaperTitleLines{\space}}
\title{\PaperTitleLines{\\}}
\author{Seungho Choi \and Sanglim Han \and Victoria Vesna \and Jihyong Oh}
\hypersetup{pdftitle={\PaperTitle},pdfauthor={Seungho Choi, Sanglim Han, Victoria Vesna, Jihyong Oh}}
\newcommand{\ProjectPageURL}{https://cmlab-korea.github.io/OTTER/}
\cmlabAuthors{Seungho Choi\textsuperscript{1,2}\quad Sanglim Han\textsuperscript{1}\quad Victoria Vesna\textsuperscript{3}\quad Jihyong Oh\textsuperscript{1,2,\textdagger}}
\newcommand{\OtterAffil}[4][1]{\mbox{\raisebox{\dimexpr-0.18\height\ifnum#3=2 -2.6pt\fi\relax}{\scalebox{#1}{\includegraphics[width=27pt,height=9pt,keepaspectratio]{\OtterArchiveRoot assets/#2}}}\enspace\textsuperscript{#3}#4}}
\cmlabAffiliations{\OtterAffil[1.5]{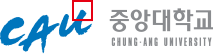}{1}{Chung-Ang University}\enspace(\OtterAffil[1.8]{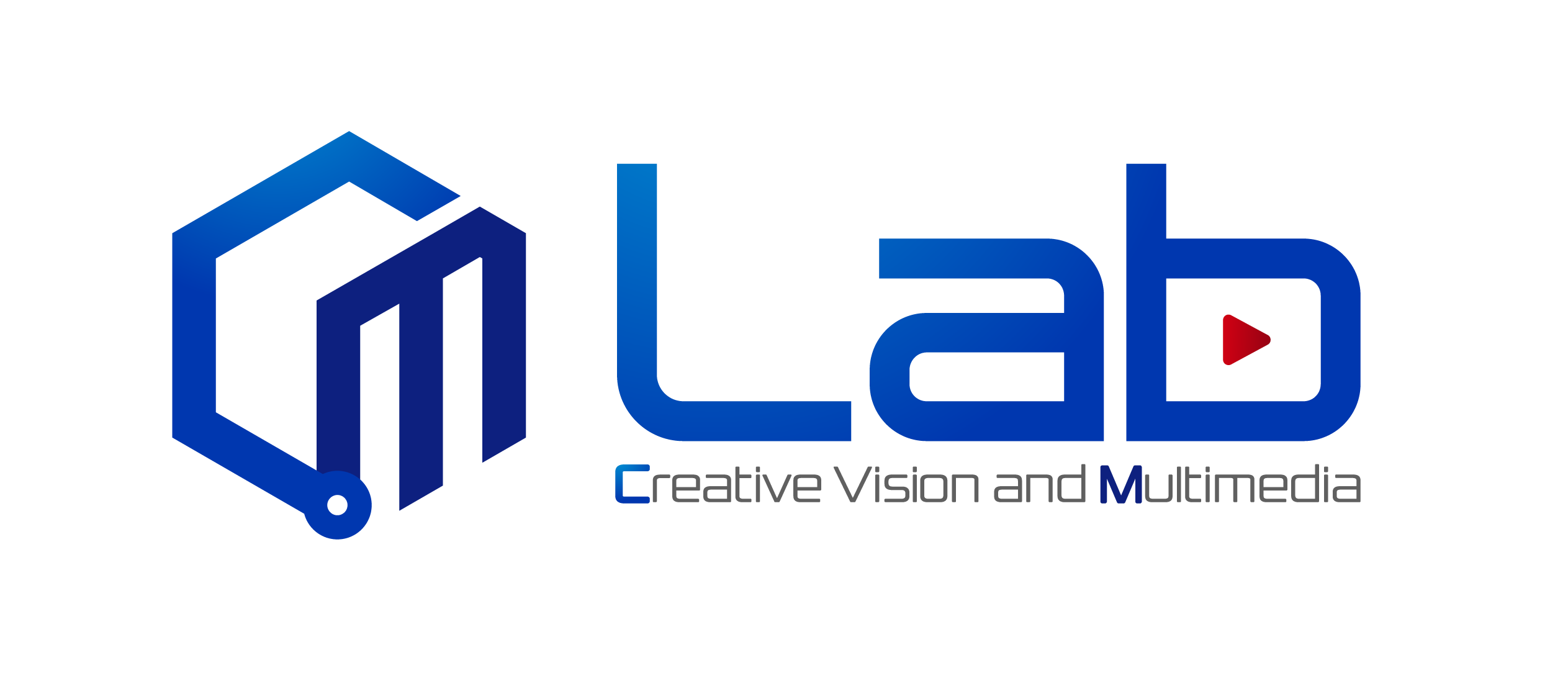}{2}{CMLab})\qquad\OtterAffil{ucla-wordmark.pdf}{3}{University of California, Los Angeles}}
\cmlabAuthorEmail{\{choiseungho1019, sanglimhan, jihyongoh\}@cau.ac.kr\qquad \href{mailto:vesna@arts.ucla.edu}{vesna@arts.ucla.edu}}
\cmlabProjectPage{https://cmlab-korea.github.io/OTTER/}
\newcommand{\ourmethod}{OTTER}

\newcommand{\cmark}{\ding{51}}
\newcommand{\xmark}{\ding{55}}
\newcommand{\tmark}{\ding{115}}

\definecolor{rrsskytext}{HTML}{0ea5e9}
\definecolor{rawpurpletext}{HTML}{9333ea}

\begin{document}
% Keep stacked full-width floats from crowding out the body text.
\setcounter{dbltopnumber}{1}
\twocolumn[{
\OtterArchiveCover
{\noindent\footnotesize This work has been submitted to the IEEE for possible publication. Copyright may be transferred without notice, after which this version may no longer be accessible.\par}\vspace{5pt}

\vspace{5pt}
{\small\setlength{\baselineskip}{10pt}\noindent\textbf{Abstract.} Conventional score distillation provides distribution-level supervision for one-step diffusion restoration but typically requires a Teacher Model and an Auxiliary Network.
Drifting removes the Auxiliary Network by estimating supervision directly from sample sets.
However, deterministic paired restoration provides only one high-quality target and one student output per input, a limitation we term \emph{conditional sample collapse}.
We address this limitation through OTTER, which introduces Nonparametric Score Distillation as a teacher-free, task- and backbone-agnostic training objective for paired image restoration.
Our central insight is that spatial recurrence within a single image pair supplies internal surrogate samples for attraction and repulsion.
We construct these sample sets from recurring high-quality residuals and corresponding student residuals relative to the low-quality input.
Attraction weights computed solely from high-quality residuals reduce sensitivity to student errors.
Regression toward the drifted target operates on the high-frequency band of decoded images, while paired reconstruction losses preserve fidelity.
In the diffusion setting, OTTER trains with a single diffusion-scale network forward pass per iteration, without a Teacher Model, Auxiliary Network, Discriminator, or alternating optimization.
Experiments on real-world image super-resolution and all-in-one image restoration demonstrate improved fidelity and perceptual similarity with U-Net, diffusion transformer, and Restormer backbones.\par}
}]
\begingroup
\renewcommand{\thefootnote}{\fnsymbol{footnote}}
\footnotetext[2]{Corresponding author.}
\endgroup
\section{Introduction}
\label{sec:intro}

\begin{figure*}[t]
  \centering
  \includegraphics[width=\linewidth]{\OtterArchiveRoot 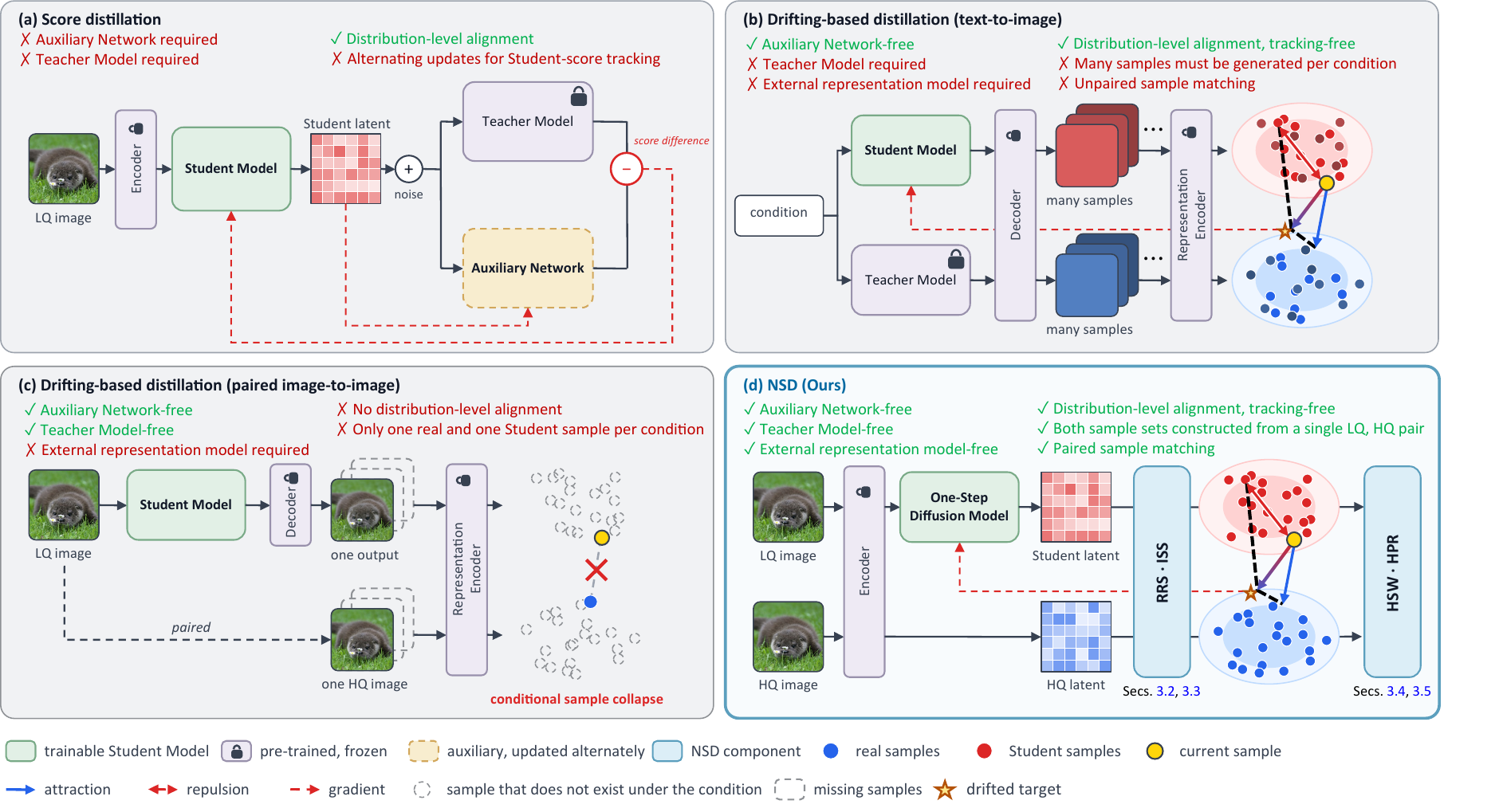}
  \caption{\textbf{One-step distillation paradigms and the challenge of
applying drifting to paired image-to-image restoration.}
(a) Score distillation aligns the Student Model with the Teacher Model
through a score difference~\cite{vsd,dmd}. Both the Teacher Model and
the Auxiliary Network evaluate the noised Student latent. The Auxiliary
Network is updated alternately with the Student Model to track the
score of its output distribution.
(b) Drifting-based distillation~\cite{drifting} removes the Auxiliary
Network by estimating the drifting field directly from real and Student
sample sets in a representation space under a shared condition
(e.g., a class label or a prompt).
When real samples per condition are scarce, as in text-to-image
generation, a frozen Teacher Model supplies the target sample
set~\cite{tfd}.
(c) In paired image-to-image restoration, each LQ image has only one
paired HQ image, and the deterministic one-step Student Model produces
only one output. Both sample sets therefore reduce to singletons,
a limitation we term \textit{conditional sample collapse}.
The dashed outlines indicate the additional samples required
but unavailable.
(d) \ourmethod{} constructs attraction and repulsion sets from
recurring internal locations through ISS (Sec.~\ref{sec:iss}),
using restoration-relevant distances in RRS (Sec.~\ref{sec:rrs}).
HSW (Sec.~\ref{sec:hsw}) computes attraction weights solely from
HQ samples to stabilize the drifted target, while HPR
(Sec.~\ref{sec:hpr}) performs regression toward this target
on the HF band in the pixel space.}
  \label{fig:concept}
\end{figure*}

Diffusion models (DMs)~\cite{ddpm,songsde} and flow-based
models~\cite{flowmatching,rectifiedflow} provide powerful generative priors
for paired image-to-image restoration, such as super-resolution, denoising, and dehazing.
In this setting, a model learns to recover a high-quality (HQ) image
from a low-quality (LQ) image using paired training examples.
For example, real-world image super-resolution
(Real-ISR) methods exploit these priors to recover realistic fine
details from degraded inputs~\cite{stablesr,diffbir,seesr,pasd,supir}.
However, iterative inference requires repeated network
evaluations for each output~\cite{ddpm}.
Accelerated sampling methods reduce the number of evaluations but
still require multiple inference steps~\cite{ddim,dpmsolver}.This computational cost motivates
one-step distillation~\cite{tsdsr,pisasr}, which compresses the iterative
restoration process into a single forward pass.

One-step distillation must recover realistic fine
details in a single forward pass.In paired image-to-image restoration, a single LQ
image can correspond to multiple plausible HQ
images~\cite{srgan,pdtradeoff}.
Pointwise regression toward the paired HQ target
can favor predictions that average over these plausible HQ images,
suppressing the high-frequency (HF) details needed for perceptually
realistic restoration~\cite{srgan,pdtradeoff}.
This averaging effect motivates
distribution-level supervision~\cite{vsd,dmd}
for one-step distillation.

Existing one-step distillation methods~\cite{progdistill,consistency,lcm,vsd,dmd,add} construct
supervision for the student model in different
ways.
These methods also differ in the additional models
required during training.
We compare these supervision strategies and training
requirements in Fig.~\ref{fig:concept} and Tab.~\ref{tab:axes}.

\begin{table}[h]
  \centering
  \caption{\textbf{Conceptual comparison of one-step
distillation families.}
\cmark: yes; \xmark: no; \tmark: varies across the cited methods.
``Auxiliary Net.'' denotes a network trained to estimate the Student
score. ``External repr.'' denotes a separate feature encoder beyond
the native latent encoder and any Teacher Model already counted.
Feature networks used solely for reconstruction losses are excluded.
The last column indicates direct applicability to the deterministic
paired image-to-image setting considered here.}
  \resizebox{\columnwidth}{!}{%
  \begin{tabular}{llccccc}
    \toprule
    & \multirow{2}{*}{Distillation family}
      & Teacher & Auxiliary & Discri- & External & Paired image- \\
    & & free & Net.\ free & minator free & repr.\ free & to-image \\
    \midrule
    (a) & Trajectory-based~\cite{progdistill,consistency,lcm}
        & \textcolor{red}{\xmark}
        & \textcolor{green}{\cmark}
        & \textcolor{green}{\cmark}
        & \textcolor{green}{\cmark}
        & \textcolor{green}{\cmark} \\
    (b) & Score distillation~\cite{vsd,dmd,dmd2,sid}
        & \textcolor{red}{\xmark}
        & \textcolor{red}{\xmark}
        & \textcolor{orange}{\tmark}
        & \textcolor{green}{\cmark}
        & \textcolor{green}{\cmark} \\
    (c) & Adversarial~\cite{add,ladd}
        & \textcolor{red}{\xmark}
        & \textcolor{green}{\cmark}
        & \textcolor{red}{\xmark}
        & \textcolor{orange}{\tmark}
        & \textcolor{green}{\cmark} \\
    (d) & Drifting-based~\cite{drifting,tfd,driftrae}
        & \textcolor{orange}{\tmark}
        & \textcolor{green}{\cmark}
        & \textcolor{green}{\cmark}
        & \textcolor{orange}{\tmark}
        & \textcolor{red}{\xmark} \\
    \midrule
    (e) & \ourmethod{} (Ours)
        & \textcolor{green}{\cmark}
        & \textcolor{green}{\cmark}
        & \textcolor{green}{\cmark}
        & \textcolor{green}{\cmark}
        & \textcolor{green}{\cmark} \\
    \bottomrule
  \end{tabular}%
  }
  \label{tab:axes}
\end{table}

\textbf{Trajectory-based distillation.}
(Tab.~\ref{tab:axes}(a))
Trajectory-based distillation trains the student model
using supervision from the ordinary differential equation (ODE)
trajectories of the teacher
model~\cite{progdistill,consistency,lcm}.
Progressive distillation uses a multi-stage curriculum to
progressively reduce the number of inference
steps~\cite{progdistill}.
Trajectory-based supervision follows individual
teacher trajectories, without explicitly measuring the discrepancy
between the student and teacher output distributions.

\noindent\textbf{Score distillation.} (Fig.~\ref{fig:concept}(a), Tab.~\ref{tab:axes}(b)).
Score distillation explicitly aligns the output
distribution of the student model with that of the teacher model.
These methods typically construct supervision from
the difference between the score functions of the teacher and student
output distributions, \textit{i.e.}, a
\textit{score difference}~\cite{vsd,dmd,sid}.

The difficulty is that the two scores are not equally
available. 
The pre-trained teacher model directly provides its
score estimate~\cite{vsd,dmd,sid}, whereas the student model
generates samples in a single forward pass without directly
estimating the score of its own output
distribution~\cite{vsd,dmd2}.
Moreover, updates to the student model change its
output distribution, so the score estimate must track this
evolving distribution~\cite{vsd,dmd2}.

To estimate the score of this evolving distribution,
existing methods introduce a trainable Auxiliary
Network~\cite{vsd,dmd,dmd2}.
Some methods~\cite{dmd,dmd2} implement the Auxiliary Network by
fine-tuning a copy of the diffusion model,
whereas others train low-rank adaptation (LoRA)
parameters~\cite{lora} attached to the frozen teacher
model~\cite{vsd,tsdsr}.
These methods alternately update the student model
and the Auxiliary Network~\cite{vsd,dmd2}.

However, the distillation objective depends on the
accuracy of the score estimate from the Auxiliary Network.
If the Auxiliary Network fails to accurately track
the evolving output distribution of the student model, its
approximation error can yield biased gradients for the student
model and lead to training instability~\cite{dmd2}.

\noindent\textbf{Adversarial distillation.}
(Tab.~\ref{tab:axes}(c))
Adversarial distillation constructs supervision
using a discriminator, at the cost of discriminator design
and objective balancing~\cite{add,ladd,arjovsky2017wgan}.
These training requirements also arise in paired
image-to-image restoration.
For example, one-step Real-ISR methods use
teacher-generated targets~\cite{sinsr},
score-based supervision~\cite{osediff,tsdsr,pisasr},
or adversarial distillation~\cite{addsr}.

\noindent\textbf{Drifting-based distillation.} (Fig.~\ref{fig:concept}(b), Tab.~\ref{tab:axes}(d))
Drifting-based distillation estimates the discrepancy
between the real distribution and the output distribution of the
student model directly from samples, without a trainable
Auxiliary Network~\cite{drifting}.
At each training step, real and student samples
obtained under the same condition form two sample sets that
represent the corresponding conditional distributions.
% When a condition has insufficient real samples,
% as in text-to-image distillation, a frozen teacher model
% generates samples that serve as substitutes for the real
% sample set~\cite{tfd}.
In settings such as text-to-image distillation,
too few real samples may be available for a given condition
to construct the required sample set.
A frozen Teacher Model generates samples under the same
condition to construct a substitute real sample set~\cite{tfd}.
The \textit{drifting field} is a vector field estimated nonparametrically from the two sets~\cite{drifting}.
At each student sample, the drifting field is the
kernel-weighted~\cite{nadaraya1964,watson1964} sum of an
\textit{attraction} toward the real samples
(Fig.~\ref{fig:concept}(b), blue arrow) and a
\textit{repulsion} from the other student samples
(Fig.~\ref{fig:concept}(b), red arrow)~\cite{drifting}.
The drifting field can be interpreted as a
nonparametric estimate of the score difference between the
two distributions~\cite{driftscore}.

% Adding the drifting field to the current student
% sample produces a \textit{drifted target}.
Adding the drifting field to the current
student sample produces a \textit{drifted target}
(Fig.~\ref{fig:concept}(b), orange star~\raisebox{-0.15em}{%
  \includegraphics[height=0.9em]{\OtterArchiveRoot 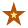}}).
The student model learns to predict this target through regression.
When the output distribution of the student model
matches the real distribution, the attraction and repulsion
terms balance, and the drifting field
vanishes~\cite{drifting,driftscore}.

Drifting-based distillation re-estimates the
drifting field directly from the two sample sets at every training
step, providing distribution-level supervision without a trainable
Auxiliary Network or a discriminator~\cite{drifting}.
Since sample distances determine the kernel weights
of attraction and repulsion, the representation should make these
distances reflect task-relevant similarities.

Subsequent methods apply drifting to one-step
distillation using the internal representations of the teacher
model~\cite{tfd} or the latents of a representation
autoencoder~\cite{driftrae}.

Existing drifting-based distillation methods~\cite{drifting,tfd,driftrae}
construct the required sample sets for class-conditional and
text-conditional generation.
In paired image-to-image restoration, however,
the condition is an individual LQ image.
The training data provide only one paired HQ image
for that condition, while the deterministic one-step student model
considered here produces only one corresponding output.
The real and student sample sets therefore each
collapse to a single sample, which we refer to as
\textit{conditional sample collapse}
(Fig.~\ref{fig:concept}(c)).
The attraction term has only the paired HQ sample, while the
repulsion term has no other student samples to act
on~\cite{drifting,nadaraya1964,watson1964}.
Applying drifting to this setting therefore
requires constructing suitable attraction and repulsion sample
sets from a single LQ-HQ pair.

To address this gap, we propose
\textbf{\ourmethod{}}, which is, to our knowledge, the \textit{first} framework
to bring drifting-based one-step distillation to paired image-to-image
restoration (Fig.~\ref{fig:concept}(d)).
\ourmethod{} is architecture-agnostic with respect
to the restoration network, which we refer to as the
\textbf{Drift-Aligned Restorer (DAR)}.
In our main setting, the DAR is a one-step diffusion
model initialized from a pre-trained generative backbone.
% The DAR takes the role of the student model in conventional
% one-step distillation, but learns without a teacher model.
% Instead, \ourmethod{} constructs a drifting field from the paired
% data and the DAR output to supervise the DAR.
The DAR takes the role of the student model
in conventional one-step distillation, but learns without
supervision from a Teacher Model.
Instead, \ourmethod{} constructs a drifted target from the
paired data and the current DAR output, and trains the DAR
through regression toward this target.
The same training objective also applies to restoration networks
trained directly on paired data without changing their architectures.
\ourmethod{} requires no trainable Auxiliary
Network~\cite{vsd,dmd,dmd2},
Discriminator~\cite{add,ladd},
or Teacher Model~\cite{progdistill,vsd,tfd}.
\ourmethod{} estimates the drifting field
nonparametrically from HQ and DAR sample
sets~\cite{drifting,driftscore} 
to provide distribution-level supervision.
% Our central insight is that spatial recurrence
% within a single LQ-HQ pair can supply internal surrogates
% for the condition-consistent samples that paired restoration
% otherwise lacks.
Our central insight is that spatial recurrence
within a single LQ-HQ pair can supply internal surrogates
for the condition-consistent samples that paired image-to-image
restoration otherwise lacks.

We realize this idea through our novel
\textbf{Internal Self-Similarity Sampling}
(ISS)~(Sec.~\ref{sec:iss}), which constructs the attraction
and repulsion sample sets from a single LQ-HQ pair.
Natural images contain recurring internal
structures, a property widely exploited by non-local
means~\cite{nlm}, collaborative filtering~\cite{bm3d}, and single-image
restoration~\cite{zontak2011,zssr}. 
For each spatial location, ISS retrieves locations
with similar HQ residuals and uses their HQ residuals and
corresponding DAR residuals to construct the attraction and
repulsion sample sets, respectively.
% ISS always retains the paired HQ sample and adds
% internal samples only from locations that satisfy its similarity
% criteria, limiting the influence of unrelated residuals on the
% drifting field.
ISS always retains the paired HQ sample
and applies validity filtering to additional locations.
The filtering prevents residuals from locations that fail
its similarity criteria from entering the drifting field.

We seek a restoration-relevant distance for
retrieving and weighting internal samples without a separate
representation encoder.
Existing drifting methods measure sample distances
using an additional feature encoder~\cite{drifting},
the internal representations of the Teacher Model~\cite{tfd},
or representation-autoencoder latents~\cite{driftrae}.
In our main diffusion setting, the variational
autoencoder (VAE) already provides latent representations.
% However, these latents are optimized for
% reconstruction~\cite{ldm}, and distances between raw latents
% can provide limited guidance for drifting-based
% training~\cite{driftrae}.
However, the VAE was originally pre-trained
for reconstruction~\cite{ldm}, and distances between raw
latents can provide limited guidance for drifting-based
training~\cite{driftrae}.
We therefore introduce our \textbf{Residual Representation Space}
(RRS)~(Sec.~\ref{sec:rrs}), which represents the HQ and DAR
latents as residuals with respect to the corresponding LQ latent.
% Using the corresponding LQ latent as a reference
% reduces the influence of shared content on sample distances,
% emphasizing the information that restoration must recover.
Using the corresponding LQ latent as a reference
reduces the influence of shared content on sample distances,
effectively emphasizing the information that restoration
must recover.
RRS thereby provides a restoration-relevant distance using the
existing VAE, without introducing a separate representation encoder.

\begin{figure}[h]
  \centering
  \includegraphics[width=\linewidth]{\OtterArchiveRoot 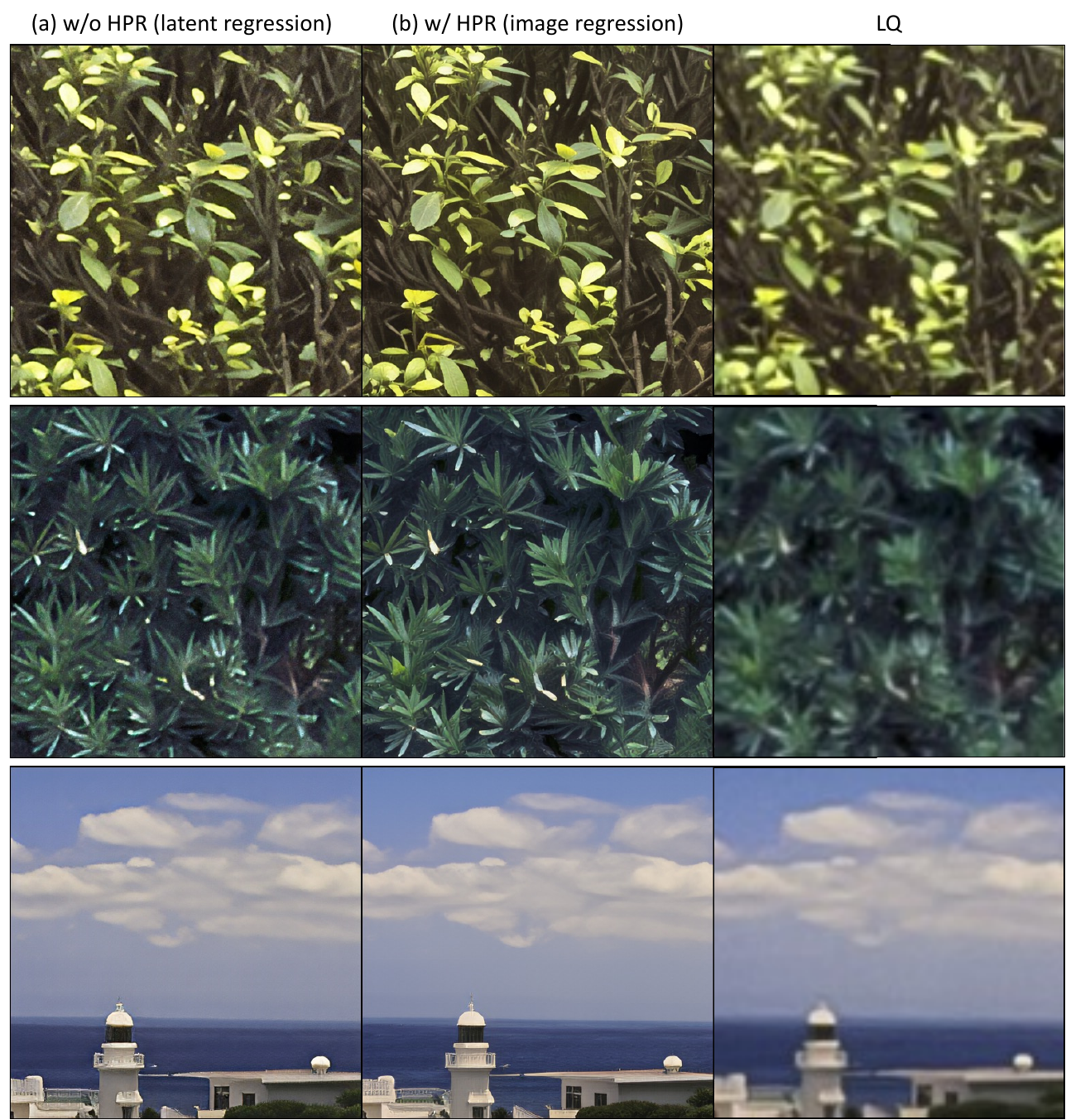}
  \caption{\textbf{Effect of the regression space.}
Both runs disable the HF split to isolate the effect of the
regression space. (a) Latent-space regression produces grid-like
artifacts. (b) Pixel-space regression preserves texture without
these artifacts. The rightmost column shows the LQ input.}
  \label{fig:latent_vs_pixel}
\end{figure}

On the other hand, attraction weights computed
from the current DAR output~\cite{drifting} can make
the supervision unreliable early in training.
Errors in the DAR output can shift these weights toward HQ
residuals that resemble the current DAR residual, reducing
the contribution of the paired HQ residual.
To address this issue, we introduce our novel
\textbf{HQ-Guided Sample Weighting}
(HSW)~(Sec.~\ref{sec:hsw}), which computes the attraction
weights from distances between HQ residuals alone.
By keeping the attraction weights independent
of the DAR output, HSW reduces the influence of DAR errors
on the drifting field and stabilizes the drifted target
to support stable training of the DAR.

Finally, we observe grid-like artifacts under direct
latent-space regression (Fig.~\ref{fig:latent_vs_pixel}).
Latent-space regression measures deviations before
decoding and does not directly measure their effects on the
restored image.
We therefore introduce our novel
\textbf{High-Frequency Pixel-Space Regression}
(HPR)~(Sec.~\ref{sec:hpr}), which measures regression toward
the drifted target on the decoded images while retaining
the drifting field in RRS.
HPR further restricts this regression to the HF
band to emphasize fine details, while paired reconstruction
losses preserve overall fidelity.

Together, these designs make drifting-based
supervision applicable to paired image-to-image restoration
without a Teacher Model, an Auxiliary Network, or a Discriminator.
The DAR learns from drifting-based supervision and paired
reconstruction losses without alternating optimization.
Experiments with U-Net and DiT backbones for one-step diffusion
distillation, together with a Restormer backbone for direct
paired restoration, demonstrate the applicability of our
backbone-agnostic training objective across these distinct
learning settings.

\section{Related Work}
\label{sec:related}

\subsection{Learning objectives for image restoration}
\label{sec:realisr}

Paired reconstruction and perceptual losses supervise
image restoration through the HQ target~\cite{srgan,lpips}, while
adversarial training adds distribution-level
supervision~\cite{srgan,realesrgan}. Diffusion-based restoration
exploits pre-trained generative
priors~\cite{stablesr,diffbir,seesr,supir}.
As discussed in Sec.~\ref{sec:intro} (Tab.~\ref{tab:axes}),
one-step diffusion restoration methods use Teacher-generated
targets, score-based supervision, or adversarial
distillation~\cite{sinsr,osediff,tsdsr,addsr}.
Drifting constructs supervision nonparametrically through
attraction and repulsion between samples~\cite{drifting,tfd,driftrae}.
\ourmethod{} adapts this principle to paired image-to-image
restoration through NSD, using internal HQ and DAR residuals
without a Teacher Model, an Auxiliary Network, or a discriminator.
We evaluate NSD alongside paired reconstruction losses with both
a latent diffusion backbone and a unmodified
Restormer backbone~\cite{Zamir2021Restormer}~(Sec.~\ref{sec:exp_main}).

\subsection{Internal statistics and frequency separation}
\label{sec:internal}

Internal statistics support non-local
filtering~\cite{nlm,bm3d,zontak2011} and image-specific
learning~\cite{zssr,kernelgan}.
\ourmethod{} uses internal recurrence to construct the HQ and DAR
residual sample sets for NSD (Sec.~\ref{sec:iss}).
Frequency separation assigns different supervision to different
image bands~\cite{fssr}.
HPR uses this principle to measure regression on the HF band of
the decoded images while retaining the drifting field in RRS
(Sec.~\ref{sec:hpr}).
\section{Method}

\subsection{Overview of \ourmethod{}}
\label{sec:overview}

\begin{figure*}[h]
\centering
\includegraphics[width=\linewidth]{\OtterArchiveRoot 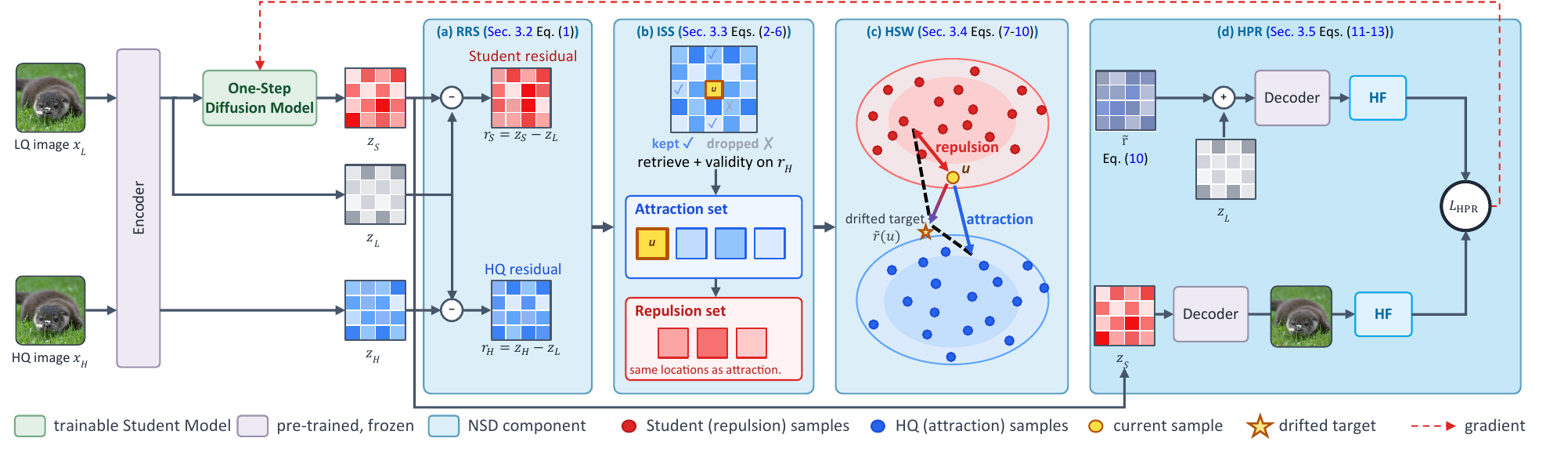}

\caption{\textbf{Overview of the \ourmethod{} framework.}
The one-step DAR $G_{\theta}$ maps the LQ latent $z_L$ to the
restored latent $z_S$. Supervision is constructed from the paired
HQ image without a Teacher Model, Auxiliary Network, or Discriminator.
(a) RRS (Sec.~\ref{sec:rrs}) represents the HQ and DAR latents as
residuals relative to the shared LQ latent, without an external
representation encoder.
(b) ISS (Sec.~\ref{sec:iss}) retrieves recurring locations in RRS
and applies validity filtering. The retained HQ and DAR residuals
form the attraction and repulsion sample sets, respectively.
(c) HSW (Sec.~\ref{sec:hsw}) computes attraction weights from
HQ-side RRS distances to reduce sensitivity to DAR errors.
The weighted sample sets define the drifting field $V(u)$
and drifted target $\tilde{r}(u)$.
(d) HPR (Sec.~\ref{sec:hpr}) decodes the DAR prediction and
drifted target and measures their regression on the
high-frequency band in the pixel space.
The VAE remains frozen throughout training.
Panels (b) and (c) show a single query $u$; the same operations
are applied to all spatial locations in parallel.}

\label{fig:pipeline}
\end{figure*}

Given an LQ image $x_L$, our framework
\ourmethod{} trains the \textbf{Drift-Aligned Restorer (DAR)}
$G_{\theta}$ to recover its paired HQ image $x_H$ in a single
forward pass. The DAR takes the role of the student model in
conventional one-step distillation, but learns without a Teacher
Model. To address \textit{conditional sample collapse}~(Sec.~\ref{sec:intro}), \ourmethod{}
constructs internal surrogate samples from spatial recurrence
within a single LQ-HQ pair. Our training objective is
backbone-agnostic; we describe the main diffusion setting,
where the DAR is initialized from a pre-trained latent diffusion
backbone~\cite{sd3}. Fig.~\ref{fig:pipeline} shows the framework.
Appendix~\ref{supp:prelim} provides the diffusion preliminaries,
the one-step restoration mapping, and the notation used below.

To measure restoration-relevant sample distances
without a separate representation encoder, our
\textbf{Residual Representation Space (RRS)}
(Sec.~\ref{sec:rrs}) represents the HQ and DAR latents as
residuals with respect to the corresponding LQ latent.
RRS reuses the existing VAE~\cite{vae} and reduces the influence
of content already present in the LQ image.
Using these distances, our
\textbf{Internal Self-Similarity Sampling (ISS)}
(Sec.~\ref{sec:iss}) retrieves locations with similar HQ
residuals and applies validity filtering.
% The retained HQ residuals and corresponding DAR residuals
% construct the attraction and repulsion sample sets, respectively,
% while ISS always retains the paired HQ sample.
We construct the attraction and repulsion
sample sets using the retained HQ residuals and corresponding
DAR residuals, respectively.

To prevent early DAR errors from changing the
attraction weights, our
\textbf{HQ-Guided Sample Weighting (HSW)}
(Sec.~\ref{sec:hsw}) computes these weights from distances
between HQ residuals alone.
These weights remain independent of the DAR output, while
the repulsion weights retain their DAR dependence.
We estimate the drifting field from the weighted sample sets
and add the field to the current DAR residual to construct
the drifted target.
HSW stabilizes this target to support
stable training of the DAR.

To directly measure the effects of regression on
the restored image, our
\textbf{High-Frequency Pixel-Space Regression (HPR)}
(Sec.~\ref{sec:hpr}) measures regression toward the drifted
target on the decoded images while retaining the drifting
field in RRS.
HPR restricts this regression to the HF band to emphasize
fine details, while paired reconstruction losses preserve
overall fidelity.
We train the DAR without a Teacher Model, an Auxiliary
Network, a Discriminator, or alternating optimization.

\subsection{Residual Representation Space (RRS)}
\label{sec:rrs}

Our \textbf{Residual Representation Space (RRS)} provides
restoration-relevant sample distances using the existing VAE,
without a separate representation encoder.
% The VAE is optimized for reconstruction~\cite{ldm},
% and distances between its raw latents can provide limited
% guidance for drifting-based training~\cite{driftrae}.
The VAE was originally pre-trained
for reconstruction~\cite{ldm}, so naively measuring sample
distances in its raw latent space can provide limited
guidance for drifting-based training~\cite{driftrae}.

Let $\mathcal{E}$ denote the frozen VAE encoder.
The LQ and HQ latents are $z_L=\mathcal{E}(x_L)$ and
$z_H=\mathcal{E}(x_H)$, and the DAR produces
$z_S=G_{\theta}(z_L)$.
We define the HQ and DAR \textit{residuals},
$r_H$ and $r_S$, as
\begin{equation}
r_H = z_H - z_L,
\qquad
r_S = z_S - z_L .
\label{eq:rrs}
\end{equation}

In Eq.~\eqref{eq:rrs}, the shared LQ latent $z_L$ provides
a reference for content already present in the input.
% Using this reference reduces the influence of shared content
% on sample distances, emphasizing the information that
% restoration must recover.
By subtracting the corresponding LQ latent,
RRS emphasizes restoration-relevant residuals in sample
distances while reducing the contribution of shared content.
We compare the RRS with raw latents in Variant (c) of
Tab.~\ref{tab:ablation} and analyze its effect on retrieval in
Sec.~\ref{sec:rrs_analysis}.

\subsection{Internal Self-Similarity Sampling (ISS) with Validity Filtering}
\label{sec:iss}

Our \textbf{Internal Self-Similarity Sampling (ISS)} addresses
conditional sample collapse by constructing multiple internal
samples from a single LQ-HQ pair.
Each LQ input provides only one paired HQ image and one
deterministic DAR output, whereas drifting requires sample sets
to estimate distribution-level supervision~\cite{drifting}.
ISS exploits recurring structures within natural images,
which provide useful samples for
restoration~\cite{nlm,bm3d,zontak2011,zssr}.
Let $p_H(u)$ denote the vectorized $3\times3$ patch centered
at $u$ in the HQ residual map $r_H$ in Eq.~\eqref{eq:rrs}.
ISS uses these patch descriptors to select locations.
At each retained location $v$, the center residual vector $r_H(v)$
provides an attraction sample, and $r_S(v)$ provides a repulsion sample.

For a query location $u$, we retrieve candidates
within a local search window $\mathcal{W}(u)$ centered at $u$.
For each candidate $v\in\mathcal{W}(u)\setminus\{u\}$,
we measure the \textit{RRS distance}
\begin{equation}
  d(u, v) = \left\lVert p_H(u) - p_H(v) \right\rVert_2 ,
\label{eq:iss_dist}
\end{equation}
and retrieve the $K$ nearest candidates together with the query itself:
\begin{equation}
  \mathcal{S}(u) = \{u\} \cup
  \operatorname*{arg\,min}_{
    \mathcal{V} \subset \mathcal{W}(u) \setminus \{u\},\; |\mathcal{V}| = K}
  \; \sum_{v \in \mathcal{V}} d(u, v) .
\label{eq:iss_topk}
\end{equation}

% We always retain $u$ because $r_H(u)$ is
% the paired HQ residual for the query.
% Since $d(u,v)$ depends only on HQ residuals, the retrieved
% locations remain independent of the current DAR output
% for a fixed training pair.
Eq.~\eqref{eq:iss_topk} retains the query location $u$
so that its paired HQ residual $r_H(u)$ is always available.
Since $d(u,v)$ depends only on HQ residuals within the same
training pair, the retrieved locations remain independent
of the current DAR output for that pair.

% Top-$K$ retrieval returns $K$ candidates even
% when no similar residuals exist, potentially introducing
% unrelated samples into the drifting field.
Naive top-$K$ retrieval returns $K$ candidates
even when no similar residuals exist, potentially introducing
unrelated samples into the drifting field.
We apply two distance criteria to filter these candidates.
First, we retain candidates whose RRS distances are
comparable to the nearest candidate distance:
\begin{equation}
  \mathcal{S}^{\dagger}(u) = \{u\} \cup
  \left\{ v \in \mathcal{S}(u) \;\middle|\;
  d(u, v) \le \gamma_{\mathrm{rel}}
  \min_{w \in \mathcal{S}(u) \setminus \{u\}} d(u, w) \right\} .
\label{eq:iss_rel}
\end{equation}

Second, we discard all additional candidates
if even the nearest candidate is too far from the query
relative to the typical RRS distance within the image:
\begin{equation}
  \min_{w \in \mathcal{S}(u) \setminus \{u\}} d(u, w)
  > \gamma_{\mathrm{abs}} \cdot
  \operatorname*{med}_{u\!,\, v}\, d(u, v)
\;\;\Longrightarrow\;\;
  \mathcal{S}^{\dagger}(u) = \{u\} .
\label{eq:iss_abs}
\end{equation}

The relative threshold $\gamma_{\mathrm{rel}}$ in
Eq.~\eqref{eq:iss_rel} and the absolute threshold
$\gamma_{\mathrm{abs}}$ in Eq.~\eqref{eq:iss_abs} are listed
in Tab.~\ref{tab:supp_hyper}.
We compute the median of the distances between queries and
candidates within each image to account for differences in
RRS distance scales across images.

From the retained locations, we construct
the \textit{attraction sample set} $\mathcal{A}(u)$ using HQ
residuals and the \textit{repulsion sample set} $\mathcal{R}(u)$
using DAR residuals at the same locations:
\begin{equation}
  \mathcal{A}(u) = \left\{ r_H(v) \right\}_{v \in \mathcal{S}^{\dagger}(u)} ,
  \qquad
  \mathcal{R}(u) = \left\{ r_S(v) \right\}_{v \in \mathcal{S}^{\dagger}(u)} .
  \label{eq:iss_sets}
\end{equation}

ISS adapts the number of internal samples
to each query while always retaining the paired HQ sample.
When no additional candidates pass the filtering,
$\mathcal{S}^{\dagger}(u)=\{u\}$, and the attraction sample set
contains only $r_H(u)$.
This fallback prevents forced inclusion of unrelated
residuals when suitable internal samples are unavailable.
We evaluate ISS and validity filtering in variants (b) and (d)
of Tab.~\ref{tab:ablation} and analyze the effective sample
size in Sec.~\ref{sec:iss_analysis}.

\subsection{HQ-Guided Sample Weighting (HSW)}
\label{sec:hsw}

Sample distances determine the attraction weights
and, consequently, which HQ samples guide the
DAR~\cite{drifting}.
In our main one-step distillation setting, the pre-trained
generative backbone must adapt to paired restoration.
During early training, errors in the DAR output can shift
the attraction weights toward HQ residuals that resemble
the current DAR residual, reducing the weight assigned to
the paired HQ residual (Fig.~\ref{fig:hsw_analysis}(b)).
We therefore propose our \textbf{HQ-Guided Sample Weighting (HSW)}, which computes
the attraction weights from distances between HQ residuals.
For a fixed training pair, these weights remain independent
of the DAR output, while the repulsion weights retain their
DAR dependence to represent the current DAR distribution.

Given the sample sets constructed by ISS
(Eq.~\eqref{eq:iss_sets}), we estimate the
\textit{drifting field} $V(u)$ using attraction toward
HQ residuals and repulsion from DAR residuals~\cite{drifting}:
\begin{equation}
\begin{aligned}
V(u)
={}&
\sum_{v \in \mathcal{S}^{\dagger}(u)}
\alpha(v)\left(r_H(v)-r_S(u)\right)
\\
&-
\sum_{v \in \mathcal{S}^{\dagger}(u)\setminus\{u\}}
\beta(v)\left(r_S(v)-r_S(u)\right),
\end{aligned}
\label{eq:field}
\end{equation}
where $\alpha(v)$ and $\beta(v)$ denote the
\textit{attraction weights} and \textit{repulsion weights}
for query $u$, respectively.
Appendix~\ref{supp:weights} gives both weight normalizations,
and Appendix~\ref{supp:weights_singleton} derives the target
when only the query is retained.

The original formulation~\cite{drifting} computes the attraction
weights using DAR-to-HQ distances,
$\alpha(v)\propto
\exp(-\lVert r_S(u)-r_H(v)\rVert_2/\tau)$,
where $\tau>0$ is the kernel bandwidth.
HSW uses the HQ residual patch distance $d(u,v)$
(Eq.~\eqref{eq:iss_dist}) to compute these weights:
\begin{equation}
\alpha(v)
\propto
\exp\left(-d(u,v)/\tau\right).
\label{eq:hsw}
\end{equation}

% Since $d(u,u)=0$, the paired HQ residual
% receives a maximum attraction weight.
Since $d(u,u)=0$ and all RRS distances
are nonnegative, the paired HQ residual receives a maximum
attraction weight under Eq.~\eqref{eq:hsw}.
HSW retains the kernel form and the DAR-dependent
repulsion in Eq.~\eqref{eq:field}.

Together, ISS and HSW determine the attraction
sample set and its weights from HQ residuals alone.
Changes in the DAR output therefore affect the drifting
field through the DAR residuals and repulsion weights,
while the attraction sample set and its weights remain fixed
for the same training pair.
Our perturbation analysis shows that HSW reduces the
sensitivity of $V(u)$ to small DAR perturbations,
particularly early in training
(Fig.~\ref{fig:hsw_analysis}(a)).
We evaluate HSW in variant (e) of Tab.~\ref{tab:ablation}
and provide further analysis in Sec.~\ref{sec:hsw_analysis}.

We construct the \textit{drifted target}
$\tilde{r}(u)$ by adding the drifting field to the DAR residual:
\begin{equation}
\tilde{r}(u) = r_S(u) + V(u).
\label{eq:target}
\end{equation}

% The targets over all latent spatial locations
% $\Omega$ constitute the \textit{drifted target map} $\tilde{r}$:
The targets over $\Omega$
(\textit{i.e.}, all positions on the two-dimensional latent grid)
constitute the \textit{drifted target map} $\tilde{r}$:
\begin{equation}
  \tilde{r} = \big[\, \tilde{r}(u) \,\big]_{u \in \Omega} ,
  \label{eq:target_map}
\end{equation}

HPR uses the drifted target map in Eq.~\eqref{eq:target_map}
to supervise the restored image in pixel space
(Sec.~\ref{sec:hpr}).

\subsection{High-Frequency Pixel-Space Regression (HPR)}
\label{sec:hpr}

Latent-space regression measures deviations
before decoding and does not directly measure their effects
on the restored image.
We observe grid-like artifacts under latent-space regression
in our comparison with pixel-space regression
(Fig.~\ref{fig:latent_vs_pixel}).
We therefore propose our \textbf{High-Frequency Pixel-Space Regression (HPR)}, which
measures regression on decoded images while retaining
the drifting field in RRS.

HPR retains the drifting field $V(u)$ and
drifted target $\tilde{r}(u)$ defined in
Eqs.~\eqref{eq:field} and~\eqref{eq:target}.
We add the LQ latent $z_L$ to the drifted target map
$\tilde{r}$ and use the frozen VAE decoder $\mathcal{D}$
to decode both the target and the DAR output:
\begin{equation}
  \tilde{x} = \mathcal{D}\!\left( z_L + \tilde{r} \right) ,
  \qquad
  x_S = \mathcal{D}\!\left( z_S \right) .
  \label{eq:hpr_decode}
\end{equation}

The full training objective consists of
the HPR loss $\mathcal{L}_{\mathrm{HPR}}$ and paired
reconstruction losses:

\begin{equation}
  \mathcal{L} =
  \mathcal{L}_{\mathrm{HPR}}
  + \lambda_{1} \left\lVert x_S - x_H \right\rVert_1
  + \lambda_{2} \, \mathcal{L}_{\mathrm{LPIPS}}\!\left( x_S, x_H \right) ,
  \label{eq:full_loss}
\end{equation}

where $\lambda_{1}$ and $\lambda_{2}$
balance the loss terms.
The paired $\ell_1$ and LPIPS~\cite{lpips} reconstruction
losses preserve overall fidelity to the HQ target.
The HPR loss restricts drifting-based supervision to the
high-frequency (HF) band to emphasize fine details.

Following frequency separation in
restoration~\cite{fssr}, we extract the HF band as
$\mathrm{HF}(x)=x-b_{\sigma}(x)$, where $b_{\sigma}$
denotes Gaussian blurring with standard deviation $\sigma$.
We define the HPR loss as

\begin{equation}
  \mathcal{L}_{\mathrm{HPR}} =
  \left\lVert
    \mathrm{HF}\!\left( x_S \right) -
    \mathrm{HF}\!\left( \tilde{x} \right)
  \right\rVert_1 .
  \label{eq:hpr_loss}
\end{equation}

% We detach the target branch and propagate
% gradients only through the DAR branch.
We detach the target branch
($\tilde{x}$ in Eq.~\eqref{eq:hpr_decode}) and propagate
gradients only through the DAR branch ($x_S$).
The VAE decoder remains frozen but differentiable,
allowing the image-space loss to update the DAR.

When $V(u)=0$ for all $u\in\Omega$,
the drifted target satisfies $\tilde{r}=r_S$.
Both branches then decode to the same image, so the HPR loss
in Eq.~\eqref{eq:hpr_loss} is zero
(Appendix~\ref{supp:weights_zero}).
We evaluate HPR in variant (a) of Tab.~\ref{tab:ablation}.
% The comparison in Fig.~\ref{fig:latent_vs_pixel}
% disables HF filtering in both runs to isolate the effect
% of the regression space (Sec.~\ref{sec:hpr_analysis}).
The run in Fig.~\ref{fig:latent_vs_pixel}(a)
measures regression between $r_S$ and $\tilde{r}$ in the
latent space, while the run in Fig.~\ref{fig:latent_vs_pixel}(b)
measures regression between $x_S$ and $\tilde{x}$ in the
pixel space.
We train both runs without the HF split to isolate the effect
of the regression space (Sec.~\ref{sec:hpr_analysis}).

The complete training objective in Eq.~\eqref{eq:full_loss}
requires no Teacher Model, Auxiliary Network, or Discriminator
and uses RRS without a separate representation encoder.
As shown in Algorithm~\ref{alg:supp_training}, each training
iteration in the diffusion setting requires one diffusion-scale
network forward pass without alternating optimization.
Appendix~\ref{supp:impl_overhead} details the remaining
training computations.
\section{Experiments}
\label{sec:exp}

\newcommand{\best}[1]{\textcolor{red}{\textbf{#1}}}
\newcommand{\snd}[1]{\textcolor{blue}{\underline{#1}}}

% Within-backbone ranking
\newcommand{\gb}[1]{\cellcolor{red!12}#1}
\newcommand{\gs}[1]{\cellcolor{blue!12}#1}

\begin{table*}[h]
  \centering
  \caption{\textbf{Quantitative comparison with state-of-the-art one-step
Real-ISR methods on RealSR, DRealSR, and DIV2K-Val.}
OTTER$_{B(\cdot)}$ denotes \ourmethod{} with backbone $B$.
Within each dataset, \protect\colorbox{red!12}{light red} and
\protect\colorbox{blue!12}{light blue} backgrounds indicate the best
and second-best results within each backbone group, respectively.
\best{Red bold} and \snd{blue underlined} text indicate the best
and second-best results across all methods, respectively.
Tied values share the same rank.
``Train Fwd.'' denotes the number of forward passes through
diffusion-scale networks per training iteration.}
  \resizebox{\linewidth}{!}{%
  \begin{tabular}{llll|ccc|ccc|c|cc|c}
    \toprule
    & Method & Venue & Backbone
      & PSNR$\uparrow$ & SSIM$\uparrow$ & FSIM$\uparrow$
      & LPIPS$\downarrow$ & DISTS$\downarrow$ & DreamSim$\downarrow$
      & pFID$\downarrow$
      & NIQE$\downarrow$ & MUSIQ$\uparrow$
      & \shortstack{Train\\Fwd.$\downarrow$} \\
    \midrule

    \multirow{7}{*}{\rotatebox{90}{RealSR}}
    & SinSR~\cite{sinsr}
      & CVPR'24 & U-Net
      & \gb{\best{26.32}} & 0.7363 & \gb{\snd{0.8565}}
      & 0.3185 & 0.2355 & 0.1144
      & 138.3 & 6.282 & 60.39 & 17 \\
    & OSEDiff~\cite{osediff}
      & NeurIPS'24 & U-Net
      & 25.14 & 0.7340 & 0.8357
      & 0.2921 & 0.2129 & 0.1188
      & 112.3 & 5.648 & 69.11 & 5 \\
    & AddSR~\cite{addsr}
      & PR'26 & U-Net
      & 22.55 & 0.6451 & 0.7740
      & 0.3823 & 0.2703 & 0.1707
      & 135.5 & 6.611 & \gb{\snd{70.68}} & 6 \\
    & PiSA-SR~\cite{pisasr}
      & CVPR'25 & U-Net
      & 25.50 & \gs{0.7481} & 0.8425
      & \gs{0.2672} & \gs{0.2044} & \gs{0.1041}
      & \gs{109.3} & \gs{5.497} & \gs{70.14} & \gs{\snd{3}} \\
    & OTTER$_{B(\text{U-Net})}$
      & N/A & U-Net
      & \gs{25.93} & \gb{\best{0.7521}} & \gs{0.8538}
      & \gb{\best{0.2576}} & \gb{\best{0.1873}} & \gb{\best{0.0941}}
      & \gb{\best{99.58}} & \gb{\best{5.004}} & 65.69 & \gb{\best{1}} \\
    \cmidrule(l){2-14}
    & TSD-SR~\cite{tsdsr}
      & CVPR'25 & DiT
      & \gs{23.40} & \gs{0.6885} & \gs{0.8325}
      & \gs{0.2805} & \gs{0.2182} & \gs{0.1051}
      & \gs{106.6} & \gb{\snd{5.094}} & \gb{\best{70.76}} & \gs{8} \\
    & OTTER$_{B(\text{DiT})}$
      & N/A & DiT
      & \gb{\snd{25.94}} & \gb{\snd{0.7516}} & \gb{\best{0.8592}}
      & \gb{\snd{0.2609}} & \gb{\snd{0.1955}} & \gb{\snd{0.0967}}
      & \gb{\snd{100.3}} & \gs{5.390} & \gs{67.10} & \gb{\best{1}} \\
    \midrule

    \multirow{7}{*}{\rotatebox{90}{DRealSR}}
    & SinSR~\cite{sinsr}
      & CVPR'24 & U-Net
      & \gs{28.36} & 0.7488 & \gs{0.8486}
      & 0.3650 & 0.2469 & 0.1427
      & 151.3 & 6.962 & 55.15 & 17 \\
    & OSEDiff~\cite{osediff}
      & NeurIPS'24 & U-Net
      & 27.91 & \gs{0.7832} & 0.8444
      & 0.2967 & \gs{0.2163} & 0.1289
      & \gs{111.9} & 6.441 & 64.66 & 5 \\
    & AddSR~\cite{addsr}
      & PR'26 & U-Net
      & 26.69 & 0.7420 & 0.8090
      & 0.3677 & 0.2625 & 0.1709
      & 133.7 & 7.655 & \gs{\snd{64.74}} & 6 \\
    & PiSA-SR~\cite{pisasr}
      & CVPR'25 & U-Net
      & 28.31 & 0.7803 & 0.8458
      & \gs{0.2959} & 0.2168 & \gs{0.1193}
      & 113.2 & \gs{\snd{6.168}} & \gb{\best{66.10}} & \gs{\snd{3}} \\
    & OTTER$_{B(\text{U-Net})}$
      & N/A & U-Net
      & \gb{\best{28.65}} & \gb{\best{0.7891}} & \gb{\snd{0.8566}}
      & \gb{0.2897} & \gb{\snd{0.1989}} & \gb{\best{0.1014}}
      & \gb{\snd{105.05}} & \gb{\best{5.556}} & 61.52 & \gb{\best{1}} \\
    \cmidrule(l){2-14}
    & TSD-SR~\cite{tsdsr}
      & CVPR'25 & DiT
      & \gs{27.45} & \gs{0.7808} & \gs{0.8410}
      & \gs{\snd{0.2861}} & \gs{0.2161} & \gs{0.1223}
      & \gs{121.7} & \gs{6.422} & \gs{60.19} & \gs{8} \\
    & OTTER$_{B(\text{DiT})}$
      & N/A & DiT
      & \gb{\snd{28.62}} & \gb{\snd{0.7841}} & \gb{\best{0.8578}}
      & \gb{\best{0.2844}} & \gb{\best{0.1978}} & \gb{\snd{0.1032}}
      & \gb{\best{102.8}} & \gb{6.202} & \gb{62.21} & \gb{\best{1}} \\
    \midrule

    \multirow{7}{*}{\rotatebox{90}{DIV2K}}
    & SinSR~\cite{sinsr}
      & CVPR'24 & U-Net
      & \gs{24.40} & 0.6014 & \gs{0.8345}
      & 0.3241 & 0.2067 & 0.1180
      & 38.7 & 6.003 & 62.82 & 17 \\
    & OSEDiff~\cite{osediff}
      & NeurIPS'24 & U-Net
      & 23.72 & \gs{0.6109} & 0.8246
      & 0.2942 & 0.1975 & 0.1185
      & 22.4 & 4.709 & 67.97 & 5 \\
    & AddSR~\cite{addsr}
      & PR'26 & U-Net
      & 22.39 & 0.5558 & 0.7831
      & 0.3819 & 0.2338 & 0.1659
      & 30.8 & 5.825 & \gs{\snd{69.11}} & 6 \\
    & PiSA-SR~\cite{pisasr}
      & CVPR'25 & U-Net
      & 23.86 & 0.6058 & 0.8289
      & \gs{0.2823} & \gs{0.1934} & \gs{0.1058}
      & \gs{22.2} & \gs{4.556} & \gb{\best{69.68}} & \gs{\snd{3}} \\
    & OTTER$_{B(\text{U-Net})}$
      & N/A & U-Net
      & \gb{\best{24.82}} & \gb{\best{0.6261}} & \gb{\snd{0.8350}}
      & \gb{\snd{0.2715}} & \gb{\best{0.1628}} & \gb{\best{0.0897}}
      & \gb{\best{17.71}} & \gb{\best{4.217}} & 65.07 & \gb{\best{1}} \\
    \cmidrule(l){2-14}
    & TSD-SR~\cite{tsdsr}
      & CVPR'25 & DiT
      & \gs{23.76} & \gs{0.6223} & \gs{0.8262}
      & \gs{0.2895} & \gs{0.1957} & \gs{0.1213}
      & \gs{26.4} & \gb{\snd{4.538}} & \gs{64.42} & \gs{8} \\
    & OTTER$_{B(\text{DiT})}$
      & N/A & DiT
      & \gb{\snd{24.63}} & \gb{\snd{0.6227}} & \gb{\best{0.8424}}
      & \gb{\best{0.2690}} & \gb{\snd{0.1650}} & \gb{\snd{0.0964}}
      & \gb{\snd{19.4}} & \gs{4.830} & \gb{67.20} & \gb{\best{1}} \\
    \bottomrule
  \end{tabular}%
  }
  \label{tab:main}
\end{table*}

\begin{table*}[t]
  \centering
  \caption{\textbf{Quantitative comparison with state-of-the-art
  all-in-one image restoration methods on CDD-11~\cite{onerestore}.}
  Results are averaged over single, double, and triple
degradation settings.
  OTTER$_{B(\cdot)}$ refers to \ourmethod{} built on backbone $B$.
  Within each degradation setting,
  \protect\colorbox{red!12}{light red} and
  \protect\colorbox{blue!12}{light blue} backgrounds indicate the best
  and second-best results within each backbone group, respectively.
  \best{Red bold} and \snd{blue underlined} text indicate the best
  and second-best results across all methods, respectively.
  Within-group background highlights are omitted for the U-Net and
  DiT groups, each of which contains a single method.}
  \label{tab:cdd11}
  \setlength{\tabcolsep}{3pt}
  \renewcommand{\arraystretch}{1.12}
  \resizebox{0.8\textwidth}{!}{%
  \begin{tabular}{llllccccccccc}
    \toprule
    & Method & Venue & Backbone
      & PSNR$\uparrow$ & SSIM$\uparrow$ & FSIM$\uparrow$
      & LPIPS$\downarrow$ & DISTS$\downarrow$ & DreamSim$\downarrow$
      & pFID$\downarrow$ & NIQE$\downarrow$ & MUSIQ$\uparrow$ \\
    \midrule

    \multirow{6}{*}{\rotatebox{90}{Single}}
     & MoCE-IR~\cite{moceir} & CVPR'25 & Restormer
      & \gs{\snd{34.92}} & \gs{\snd{0.9506}} & \gs{\snd{0.9785}}
      & \gs{\snd{0.0545}} & \gs{\snd{0.0524}} & \gs{\snd{0.0148}}
      & \gs{\snd{9.24}} & 5.586 & \gs{\snd{69.79}} \\
     & BaryIR~\cite{baryir} & TPAMI'26 & Restormer
      & 22.16 & 0.8467 & 0.9192
      & 0.2111 & 0.1515 & 0.0916
      & 31.92 & 5.494 & 64.51 \\
     & MIRAGE~\cite{mirage} & ICLR'26 & Restormer
      & 22.48 & 0.8574 & 0.9214
      & 0.1938 & 0.1409 & 0.0837
      & 26.27 & \gs{5.429} & 65.56 \\
     & \ourmethod{}$_{B(\text{Restormer})}$ & N/A & Restormer
      & \gb{\best{36.08}} & \gb{\best{0.9517}} & \gb{\best{0.9818}}
      & \gb{\best{0.0286}} & \gb{\best{0.0265}} & \gb{\best{0.0066}}
      & \gb{\best{4.61}} & \gb{5.386} & \gb{\best{70.11}} \\
    \cmidrule(l){2-13}
     & \ourmethod{}$_{B(\text{U-Net})}$ & N/A & U-Net
      & 23.23 & 0.6100 & 0.8902
      & 0.1531 & 0.0823 & 0.0333
      & 18.83 & \snd{4.126} & 68.07 \\
    \cmidrule(l){2-13}
     & \ourmethod{}$_{B(\text{DiT})}$ & N/A & DiT
      & 25.02 & 0.7301 & 0.9415
      & 0.0976 & 0.0662 & 0.0198
      & 15.19 & \best{4.081} & 69.47 \\
    \midrule

    \multirow{6}{*}{\rotatebox{90}{Double}}
     & MoCE-IR~\cite{moceir} & CVPR'25 & Restormer
      & \gs{\snd{30.06}} & \gs{\snd{0.8892}} & \gs{\snd{0.9568}}
      & \gs{\snd{0.1214}} & \gs{0.0993} & \gs{\snd{0.0322}}
      & \gs{21.43} & \gs{5.162} & \gs{68.53} \\
     & BaryIR~\cite{baryir} & TPAMI'26 & Restormer
      & 17.99 & 0.6821 & 0.8219
      & 0.4663 & 0.2739 & 0.2374
      & 82.91 & 6.360 & 57.05 \\
     & MIRAGE~\cite{mirage} & ICLR'26 & Restormer
      & 17.92 & 0.7080 & 0.8352
      & 0.4124 & 0.2480 & 0.2134
      & 65.87 & 5.918 & 60.14 \\
     & \ourmethod{}$_{B(\text{Restormer})}$ & N/A & Restormer
      & \gb{\best{30.83}} & \gb{\best{0.8910}} & \gb{\best{0.9625}}
      & \gb{\best{0.0641}} & \gb{\best{0.0504}} & \gb{\best{0.0147}}
      & \gb{\best{9.65}} & \gb{4.928} & \gb{\best{69.73}} \\
    \cmidrule(l){2-13}
     & \ourmethod{}$_{B(\text{U-Net})}$ & N/A & U-Net
      & 19.12 & 0.5711 & 0.8585
      & 0.2237 & 0.1256 & 0.0905
      & 28.44 & \snd{4.074} & 66.46 \\
    \cmidrule(l){2-13}
     & \ourmethod{}$_{B(\text{DiT})}$ & N/A & DiT
      & 23.89 & 0.6842 & 0.9146
      & 0.1426 & \snd{0.0899} & 0.0385
      & \snd{19.64} & \best{4.007} & \snd{68.56} \\
    \midrule

    \multirow{6}{*}{\rotatebox{90}{Triple}}
     & MoCE-IR~\cite{moceir} & CVPR'25 & Restormer
      & \gs{\snd{27.38}} & \gs{\snd{0.8186}} & \gs{\snd{0.9344}}
      & \gs{0.1978} & \gs{0.1479} & \gs{\snd{0.0533}}
      & \gs{43.17} & \gs{4.683} & \gs{66.55} \\
     & BaryIR~\cite{baryir} & TPAMI'26 & Restormer
      & 17.66 & 0.5499 & 0.7231
      & 0.6895 & 0.3717 & 0.3890
      & 142.20 & 7.079 & 44.39 \\
     & MIRAGE~\cite{mirage} & ICLR'26 & Restormer
      & 17.47 & 0.5967 & 0.7547
      & 0.6027 & 0.3406 & 0.3532
      & 115.70 & 5.827 & 52.49 \\
     & \ourmethod{}$_{B(\text{Restormer})}$ & N/A & Restormer
      & \gb{\best{27.43}} & \gb{\best{0.8206}} & \gb{\best{0.9415}}
      & \gb{\best{0.1068}} & \gb{\best{0.0760}} & \gb{\best{0.0245}}
      & \gb{\best{18.58}} & \gb{4.371} & \gb{\best{69.15}} \\
    \cmidrule(l){2-13}
     & \ourmethod{}$_{B(\text{U-Net})}$ & N/A & U-Net
      & 18.09 & 0.5514 & 0.8214
      & 0.3075 & 0.1684 & 0.1768
      & 38.73 & \snd{3.991} & 65.64 \\
    \cmidrule(l){2-13}
     & \ourmethod{}$_{B(\text{DiT})}$ & N/A & DiT
      & 23.06 & 0.6392 & 0.8889
      & \snd{0.1928} & \snd{0.1124} & 0.0610
      & \snd{28.81} & \best{3.891} & \snd{67.17} \\
    \bottomrule
  \end{tabular}%
  }
\end{table*}

\subsection{Experimental Settings}
\label{sec:exp_settings}

\noindent\textbf{Training and test datasets.}
% Following SeeSR~\cite{seesr} and TSD-SR~\cite{tsdsr},
% we use LSDIR~\cite{lsdir} together with DIV2K~\cite{div2k} and
% Flickr2K~\cite{flickr2k} for training, and synthesize LQ-HQ pairs with the
% degradation pipeline of Real-ESRGAN~\cite{realesrgan}. 
For Real-ISR, following SeeSR~\cite{seesr} and
TSD-SR~\cite{tsdsr}, we use LSDIR~\cite{lsdir} together with
DIV2K~\cite{div2k} and Flickr2K~\cite{flickr2k} for training,
and synthesize LQ-HQ pairs with the degradation pipeline of
Real-ESRGAN~\cite{realesrgan}.
We evaluate all methods on RealSR~\cite{realsrset}, DRealSR~\cite{drealsr}, and DIV2K-Val~\cite{div2k}.
% For all-in-one image restoration, we train and evaluate on
% CDD-11~\cite{onerestore}, which covers four degradation factors and their
% composites over eleven settings.
For all-in-one image restoration,
we train and evaluate on CDD-11~\cite{onerestore} for all-in-one
image restoration. CDD-11 covers four degradation factors and
their composites over eleven settings, which we group into
single, double, and triple degradations.

\noindent\textbf{Evaluation metrics.}
We report the same full-reference and no-reference metrics for both tasks.
PSNR, SSIM~\cite{ssim}, and FSIM~\cite{fsim} measure fidelity to the paired HQ image.
LPIPS~\cite{lpips}, DISTS~\cite{dists}, and
DreamSim~\cite{dreamsim} measure perceptual similarity to the paired HQ image.
Patch FID (pFID)~\cite{supir} measures the distance between
the distributions of $256\times256$ patches from the restored
and HQ images. 
The no-reference metrics are NIQE~\cite{niqe} and
MUSIQ~\cite{musiq}.

\noindent\textbf{Compared methods.}
For one-step diffusion-based Real-ISR,
we compare with the U-Net-based methods SinSR~\cite{sinsr},
OSEDiff~\cite{osediff}, AddSR~\cite{addsr}, and
PiSA-SR~\cite{pisasr}, and the DiT-based method
TSD-SR~\cite{tsdsr}.
All results are obtained
with the official code and weights under the same evaluation
protocol, following SeeSR~\cite{seesr} and
TSD-SR~\cite{tsdsr}.
On CDD-11, we compare with all-in-one restoration
methods: MoCE-IR~\cite{moceir}, BaryIR~\cite{baryir}, and
MIRAGE~\cite{mirage}.

Please note that these methods employ specialized
architectural designs, such as complexity experts with adaptive
routing in MoCE-IR~\cite{moceir}, a learnable Wasserstein
barycenter mapping module in BaryIR~\cite{baryir}, and
channel-wise decomposition into CNN, attention, and MLP
branches in MIRAGE~\cite{mirage}.
In contrast, we adopt the original Restormer~\cite{Zamir2021Restormer}
% TODO: Restormer 원논문의 기존 citation key로 교체
without any architectural modification.
% This constraint requires our training objective to handle
% diverse degradations using the original backbone alone.
% Despite this architectural constraint, our Restormer variant
% outperforms these specialized methods on all six full-reference
% metrics, pFID, and MUSIQ across single, double, and triple
% degradations on CDD-11 (Tab.~\ref{tab:cdd11}), demonstrating
% the effectiveness of our training objective with an
% unmodified restoration backbone.

\noindent\textbf{Implementation details.}
% We use SD3~\cite{sd3} as the latent diffusion
% backbone and train the DAR $G_{\theta}$ by attaching trainable
% LoRA~\cite{lora} layers, following the configuration of
% TSD-SR~\cite{tsdsr}. The input timestep of the backbone is fixed
% to the same value as in TSD-SR.
% The VAE encoder and decoder are frozen. 
In our main one-step diffusion distillation
setting, we instantiate the DAR $G_{\theta}$ with pretrained
U-Net and DiT backbones and train it by attaching trainable
LoRA~\cite{lora} layers.
For direct paired restoration, we instantiate
the DAR with a Restormer backbone and train it from scratch
on CDD-11 using our training objective.
Across backbones, we adjust only the learning rate
and keep all hyperparameters of our method identical,
including the search window and patch size of ISS,
the validity thresholds, the bandwidth of HSW,
the HF split of HPR, and the loss weights.
This shared configuration requires no backbone-specific
tuning of our method's hyperparameters.
Tabs.~\ref{tab:supp_realisr_training} and~\ref{tab:supp_train_config}
list the training settings for Real-ISR and CDD-11, respectively.
Tab.~\ref{tab:supp_hyper} lists the shared NSD hyperparameters,
and Appendix~\ref{supp:impl_restormer} describes training with Restormer.

% \begin{figure*}[h]
% \centering
% \includegraphics[width=\textwidth]{\OtterArchiveRoot figs/qualitative_sr.pdf}
% \caption{Qualitative comparison with one-step Real-ISR methods~\cite{sinsr,addsr,tsdsr,pisasr} on
% RealSR~\cite{realsrset}, DRealSR~\cite{drealsr}, and
% DIV2K-Val~\cite{div2k}.}
% \label{fig:qualitative_sr}
% \end{figure*}
\begin{figure*}[h]
\centering
\begin{subfigure}[t]{0.505\textwidth}\centering\caption{Real-ISR comparison with one-step methods%
~\cite{sinsr,addsr,tsdsr,pisasr} on RealSR~\cite{realsrset},
DRealSR~\cite{drealsr}, and DIV2K-Val~\cite{div2k}.}
\label{fig:qualitative_sr}
  \includegraphics[width=\linewidth]{\OtterArchiveRoot 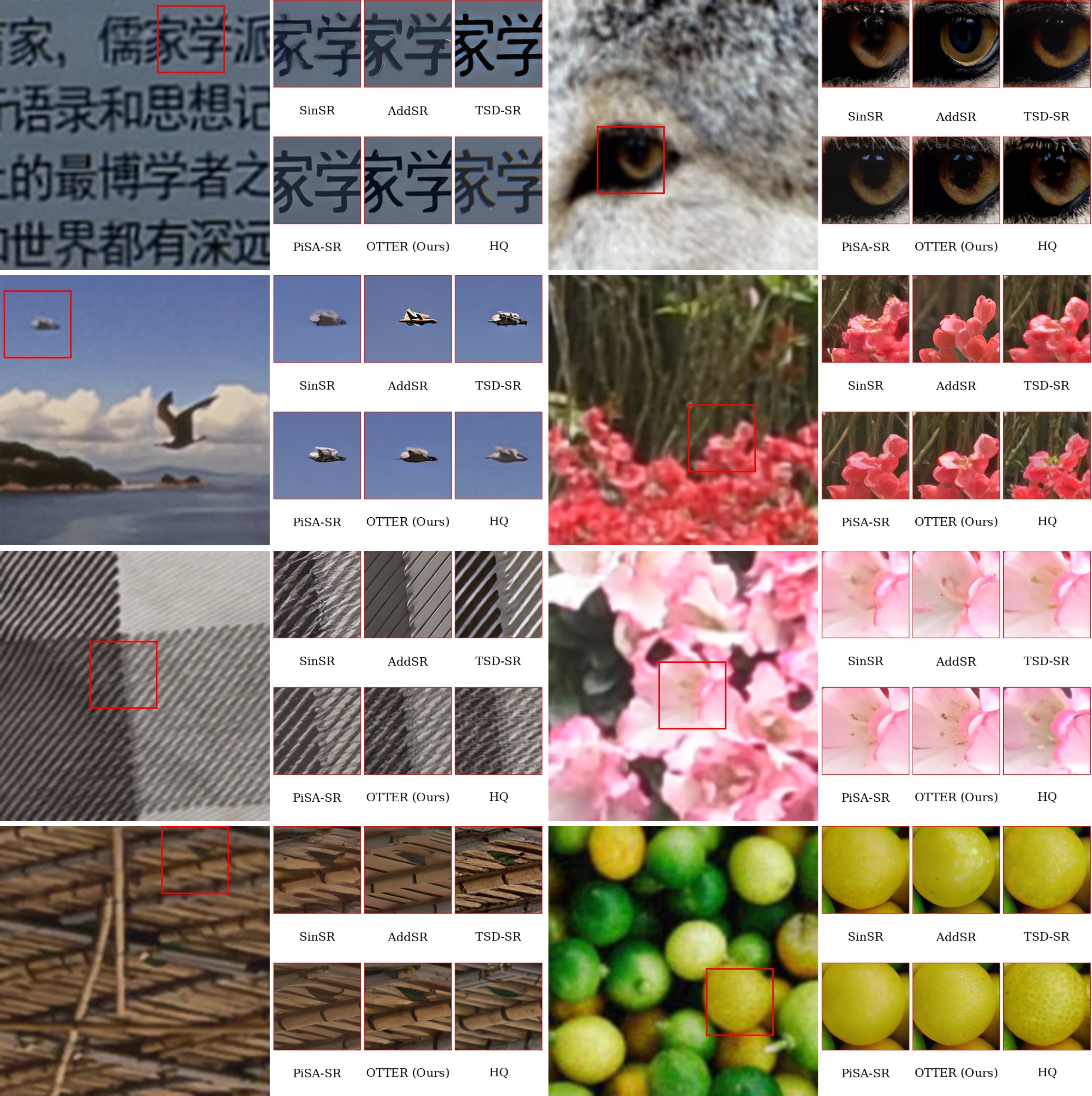}%
\end{subfigure}
\hfill
\begin{subfigure}[t]{0.475\textwidth}\centering\caption{All-in-one image restoration comparison
on CDD-11~\cite{onerestore}.}
\label{fig:qualitative_restor}
  \includegraphics[width=\linewidth]{\OtterArchiveRoot 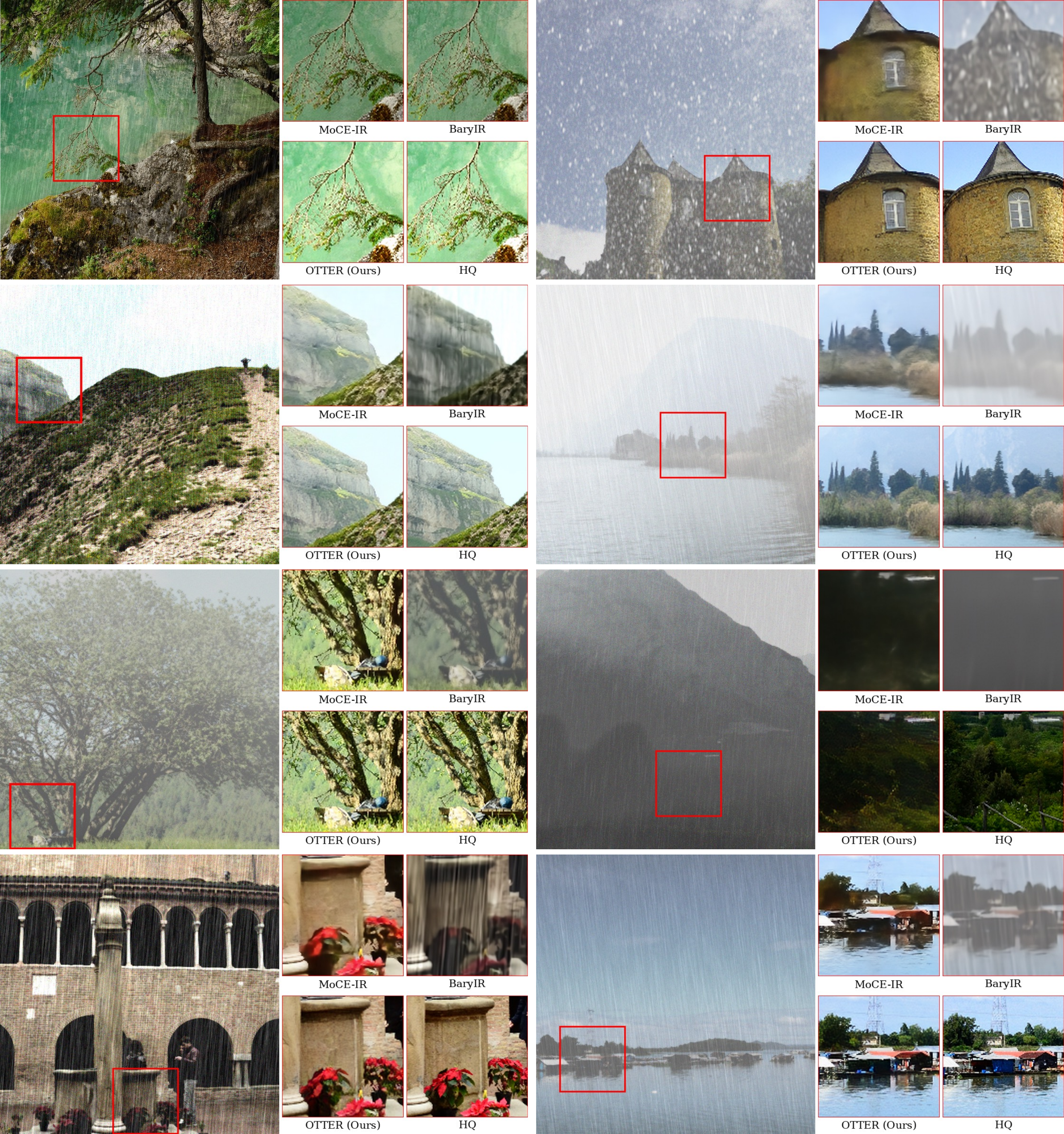}%
\end{subfigure}
\caption{Qualitative comparisons for Real-ISR and all-in-one
image restoration. Red boxes indicate the regions enlarged
for comparison.}
\label{fig:qualitative}
\end{figure*}

% \begin{figure*}[h]
% \centering
% % \includegraphics[width=\textwidth]{\OtterArchiveRoot figs/qualitative_cdd.pdf}
% \caption{Qualitative comparison with all-in-one restoration methods~\cite{moceir,baryir} on
% CDD-11~\cite{onerestore}.}
% \label{fig:qualitative_cdd}
% \end{figure*}

\subsection{Comparison with State-of-the-Art Methods}
\label{sec:exp_main}

\noindent\textbf{Quantitative comparison.}
% Tab.~\ref{tab:main} reports the quantitative comparison
% on RealSR, DRealSR, and DIV2K-Val. On RealSR, \ourmethod{} attains
% the best SSIM and FSIM, improves LPIPS, DISTS, and DreamSim over
% the strongest competitor in each metric, and obtains the best pFID.
% It ranks second in PSNR, behind SinSR. On DRealSR and DIV2K-Val,
% \ourmethod{} achieves the best scores on all six full-reference
% metrics and pFID. These improvements also extend to DreamSim,
% which is not included in our training objective.
As shown in Tab.~\ref{tab:main}, \ourmethod{}
improves both the U-Net and DiT backbones for one-step Real-ISR.
Within the U-Net group, our variant achieves the best SSIM, LPIPS,
DISTS, DreamSim, and pFID on all three benchmarks, and the best
PSNR and FSIM on DRealSR and DIV2K-Val.
Within the DiT group, our variant outperforms TSD-SR on all six
full-reference metrics and pFID on all three benchmarks.
Across all compared methods, one of the two \ourmethod{} variants
achieves the best score on every full-reference metric and pFID
on all three benchmarks, except for PSNR on RealSR, where SinSR
is the best.
Note that these gains include DreamSim, which is not included in
our training objective.

On the no-reference metrics, our U-Net variant~(OTTER$_{B(\text{U-Net})}$)
achieves the best NIQE across all compared methods on all three
benchmarks, and our DiT variant~(OTTER$_{B(\text{DiT})}$) achieves higher MUSIQ than TSD-SR
on DRealSR and DIV2K-Val. On MUSIQ, both variants are lower than
PiSA-SR and AddSR. However, a higher MUSIQ is accompanied by lower
fidelity in both backbone groups. In the U-Net group, AddSR obtains
the second best MUSIQ on all three benchmarks but shows the lowest
PSNR and the highest LPIPS among all compared methods. In the DiT
group, TSD-SR obtains the best MUSIQ on RealSR, while its PSNR is
considerably lower than that of our DiT variant. These results show
that, with both backbones, \ourmethod{} obtains competitive
no-reference quality without sacrificing fidelity to the paired
HQ image.

Finally, \ourmethod{} \textit{only} requires a single forward
pass through a diffusion-scale network per training iteration,
compared to 3 for PiSA-SR and 8 for TSD-SR
(Tab.~\ref{tab:main}). This is the fewest diffusion-scale forward
passes among the compared methods.

\noindent\textbf{All-in-one image restoration.}
We further evaluate \ourmethod{} on
CDD-11~\cite{onerestore} to examine two questions:
(i) whether \ourmethod{} generalizes from Real-ISR to a different
restoration task, and (ii) whether our training objective remains
effective outside one-step diffusion distillation.
For (i), we train our U-Net and DiT variants on CDD-11 in the same
one-step diffusion distillation setting as in Real-ISR.
For (ii), we train our Restormer variant from scratch by direct
paired restoration (Sec.~\ref{sec:exp_settings}).

As shown in Tab.~\ref{tab:cdd11}, our Restormer
variant achieves the best score across all compared methods on all
six full-reference metrics, pFID, and MUSIQ for single, double, and
triple degradations, and the best NIQE within the Restormer group.
Compared with MoCE-IR, which is the strongest compared method, our
Restormer variant reduces LPIPS, DISTS, DreamSim, and pFID by
approximately half in all three settings, while also improving
PSNR, SSIM, and FSIM.
Note that our Restormer variant adopts the original Restormer
architecture, whereas the compared methods introduce additional
architectural designs.
These results show that our training objective is effective for
direct paired restoration with a conventional backbone.

In contrast, our U-Net and DiT variants obtain lower PSNR
and SSIM than our Restormer variant and MoCE-IR across all
three degradation groups.
The strong fidelity of our Restormer variant under the same
training objective demonstrates that NSD can support accurate
paired reconstruction.
The gap observed with the diffusion variants therefore
motivates examining how the pre-trained generative backbones
balance detail generation and preservation of the input content.
For mildly degraded inputs, much of the underlying structure
remains available, making preservation of reliable input content
particularly important.
Generating plausible details can reduce fidelity when those
details depart from the content of the paired HQ image.
Hu et al.~\cite{ICLR2026_baa66ca8} suggest that diffusion-based
restorers may interpret mildly degraded inputs as severely
degraded and consequently generate details inconsistent with
the LQ image in all-in-one restoration.
This account provides a possible explanation for the larger
fidelity gap in our single-degradation results, where preserving
input-specific structure may be more beneficial than generating
additional detail.

The relative performance of our DiT variant improves under
composite degradations.
From single to triple degradations, its PSNR gap relative to
MoCE-IR narrows.
This trend is accompanied by lower DISTS and pFID for double
and triple degradations, together with lower LPIPS for triple
degradations.
The improvements in DISTS and LPIPS indicate greater perceptual
similarity to the paired HQ images, while the lower pFID
indicates closer alignment with the distribution of HQ patches.
These results provide complementary evidence that the DiT
variant becomes more competitive with MoCE-IR in recovering
perceptually relevant content under composite degradations,
even though its PSNR and SSIM remain lower.
Both diffusion variants also achieve the best and second-best
NIQE across all compared methods in every degradation group.
The NIQE results indicate favorable no-reference quality,
while the paired metrics reveal the remaining differences
in fidelity to the HQ images.
Overall, the shared NSD objective supports distinct restoration
profiles: our Restormer variant provides the strongest paired
fidelity, while our DiT variant becomes more competitive with
MoCE-IR on perceptual similarity and patch-distribution metrics
as the number of degradation factors increases.

\noindent\textbf{Qualitative comparison.}
Fig.~\ref{fig:qualitative_sr} presents visual comparisons on the three
benchmarks. On repetitive textures such as the knit pattern, \ourmethod{}
restores the recurring structure with consistent fine details, while the
compared methods blur the pattern or break its regularity. On fine
structures such as the petals and the eye, \ourmethod{} recovers sharp HF
details without the over-smoothing of regression-oriented outputs or the
artifacts introduced by the compared one-step methods. Under severe
degradations, as in the ornament example, several compared methods
hallucinate structures that deviate from the underlying content, whereas
\ourmethod{} restores fine details that remain faithful to the HQ image.
These observations are consistent with the quantitative results, where
\ourmethod{} improves perceptual similarity to the HQ image alongside
fidelity.
Figs.~\ref{fig:supp_realisr} and~\ref{fig:supp_cdd} provide
additional visual comparisons for Real-ISR and CDD-11, respectively.
Fig.~\ref{fig:supp_failure} shows challenging examples from both tasks.

\noindent\textbf{User study.}
With separate consent, 33 volunteers rate 20 Real-ISR and 20 CDD-11
examples for reference fidelity and perceptual quality on five-point
scales. We collect no identifying information and did not obtain prior
IRB review. Our U-Net and Restormer variants receive the highest mean
ratings on both criteria for their respective tasks
(Tab.~\ref{tab:user_study}). Appendix~\ref{supp:user_protocol} details
the protocol; Tab.~\ref{tab:supp_user_distribution} reports score
distributions.

\begin{table}[h]
\centering
\caption{Mean user ratings on a five-point scale.
Fid.: fidelity; PQ: perceptual quality.
Higher is better; the best scores are bold.}
\label{tab:user_study}
\footnotesize
\setlength{\tabcolsep}{3pt}
\begin{tabular}{lcc@{\hspace{10pt}}lcc}
\toprule
\multicolumn{3}{c}{Real-ISR} &
\multicolumn{3}{c}{CDD-11} \\
\cmidrule(lr){1-3}\cmidrule(lr){4-6}
Method & Fid.$\uparrow$ & PQ$\uparrow$ &
Method & Fid.$\uparrow$ & PQ$\uparrow$ \\
\midrule
AddSR  & 2.45 & 2.97 & BaryIR  & 2.44 & 2.46 \\
TSD-SR & 3.41 & 3.30 & MIRAGE  & 2.43 & 2.49 \\
PiSA-SR & 3.74 & 3.93 & MoCE-IR & 3.90 & 3.83 \\
\ourmethod{} & \textbf{4.04} & \textbf{4.06} &
\ourmethod{} & \textbf{4.71} & \textbf{4.71} \\
\bottomrule
\end{tabular}
\end{table}

\subsection{Ablation Study and Analysis}
\label{sec:ablation}
\label{sec:analysis}

Tab.~\ref{tab:ablation} reports component ablations
of \ourmethod{} on RealSR. Every variant underperforms the full
model across all reported metrics. We examine these performance
changes alongside analyses of each component.
  
\begin{table}[h]
  \centering
  \caption{\textbf{Component ablation on RealSR.}
Each variant removes or replaces one design choice while keeping
the remaining settings fixed. Variant (c) uses raw latents without
residual subtraction. The best and second-best results are
highlighted in \best{red} and \textcolor{blue}{\underline{blue}},
respectively.}
  \resizebox{\columnwidth}{!}{%
  \begin{tabular}{lcc|cc|c}
    \toprule
    Variant & PSNR$\uparrow$ & SSIM$\uparrow$
            & LPIPS$\downarrow$ & DISTS$\downarrow$ & pFID$\downarrow$ \\
    \midrule
    (a) w/o HPR (latent regression)           & 21.74 & 0.6228 & 0.3849 & 0.2843 & 170.3 \\
    (b) w/o ISS ($\mathcal{S}(u) = \{u\}$)    & 22.33 & 0.6423 & 0.3684 & 0.2676 & 162.3 \\
    (c) w/o RRS (raw latents)          & 23.17 & 0.6524 & 0.3551 & 0.2649 & 148.8 \\
    (d) w/o validity filtering                & 24.39 & 0.6845 & 0.3067 & 0.2161 & 121.7 \\
    (e) w/o HSW (weights from $r_S(u)$)       & \snd{25.16} & \snd{0.7180} & \snd{0.2838} & \snd{0.2058} & \snd{102.8} \\
    OTTER (full)                              & \best{25.94} & \best{0.7516} & \best{0.2609} & \best{0.1955} & \best{100.3} \\
    \bottomrule
  \end{tabular}}
  \label{tab:ablation}
\end{table}

\subsubsection{RRS: the representation space.}
\label{sec:rrs_analysis}

\begin{figure}[h]
  \centering
  \includegraphics[width=\columnwidth]{\OtterArchiveRoot 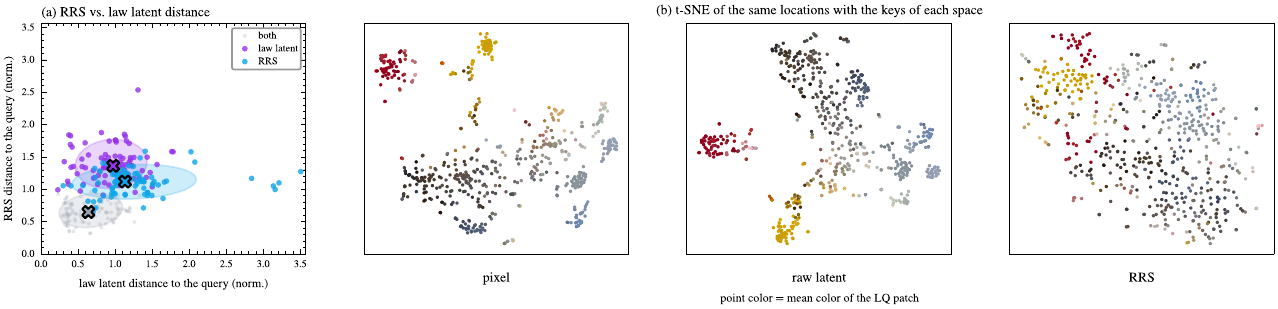}
  \caption{\textbf{What the RRS changes in the retrieval.} (a) Each
retrieved candidate, plotted by its raw latent distance and its RRS
distance $d$ to the query (Eq.~\eqref{eq:iss_dist}). RRS-only candidates
(\textcolor{rrsskytext}{\textbf{sky blue}}) are far in the raw latent
space but close in the RRS; raw-only candidates
(\textcolor{rawpurpletext}{\textbf{purple}}) show the opposite pattern.
(b) t-SNE embeddings of the same locations from their pixel, raw
latent, and RRS keys (Eq.~\eqref{eq:rrs}), each point colored by the mean color of its LQ
patch. Locations cluster by color in the pixel and raw
latent spaces; this color-based clustering is less pronounced
in the RRS.}
  \label{fig:retrieval}
\end{figure}

Variant (c) in Tab.~\ref{tab:ablation} replaces
the RRS with raw latents without residual subtraction and performs
worse across all reported metrics. To examine how the RRS
(Sec.~\ref{sec:rrs}) affects retrieval, we sample query
locations from the training sets~\cite{lsdir,div2k,flickr2k} and retrieve
the $K$ nearest candidates for each query twice, using either the raw
latent distance or the RRS distance $d$ (Eq.~\eqref{eq:iss_dist}).
Fig.~\ref{fig:retrieval}(a) plots each retrieved candidate according to
its raw latent and RRS distances to the query. Since the two retrievals
select mostly the same candidates, we focus on those selected by only
one. RRS-only candidates are far from the query in the raw latent space
but close in the RRS. Raw-only candidates exhibit the opposite pattern.
These results show that RRS can retrieve similar restoration residuals
from locations that raw latent retrieval overlooks.

Fig.~\ref{fig:retrieval}(b) further examines
the three representation spaces. We construct separate
t-SNE embeddings of the same sampled locations using their pixel, raw
latent, and RRS keys, and color each point by the mean color of its LQ
patch. In the pixel and raw latent spaces, locations cluster according to
color. This color-based clustering is less pronounced in the RRS,
consistent with the residual construction in Eq.~\eqref{eq:rrs}
reducing the influence of the shared LQ content.
Together with the ablation results, these observations support
using restoration residuals to define the sample distances.

\subsubsection{ISS: adaptivity of the sample sets.}
\label{sec:iss_analysis}

\begin{figure}[h]
  \centering
  \includegraphics[width=\columnwidth]{\OtterArchiveRoot 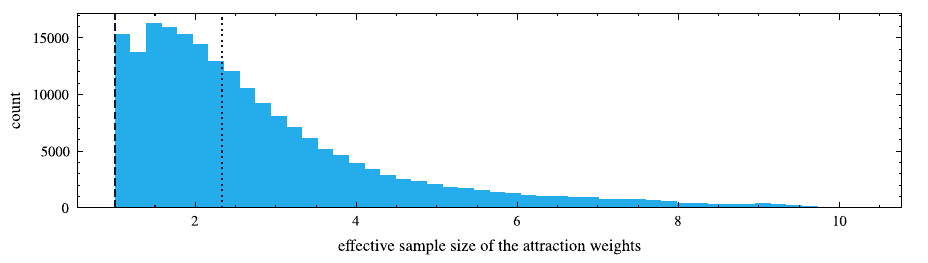}
  \caption{\textbf{Effective sample size of the
attraction weights $\alpha$.}
The distribution of
$\mathrm{ESS}(u)=
1/\sum_{v\in\mathcal{S}^{\dagger}(u)}\alpha(v)^2$
for normalized attraction weights over query locations sampled
from the training sets. The dashed line marks $\mathrm{ESS}=1$,
where a single attraction sample carries all the weight, and
the dotted line marks the median. Most locations have an ESS
between two and three. The upper tail extends toward the maximum
of $K+1$, while values near one indicate concentration on a
single sample, including singleton fallback cases.}
  \label{fig:iss_ess}
\end{figure}

Variants (b) and (d) in Tab.~\ref{tab:ablation}
show the roles of internal sampling and validity filtering.
Without ISS, as in Variant (b), $\mathcal{S}(u)=\{u\}$, so the
attraction and repulsion sets $\mathcal{A}(u)$ and $\mathcal{R}(u)$
in Eq.~\eqref{eq:iss_sets} both become singletons at the paired
location. This setting reproduces conditional sample collapse and
produces the second-largest performance loss. Without validity
filtering, as in Variant (d), every candidate in $\mathcal{S}(u)$
(Eq.~\eqref{eq:iss_topk}) is retained without applying
Eqs.~\eqref{eq:iss_rel} and~\eqref{eq:iss_abs}.
The performance losses of both variants support constructing
multiple internal samples and filtering them by their similarity
to the query in RRS (Sec.~\ref{sec:iss}).

To determine how many attraction samples actually
contribute, we use the effective sample size (ESS)~\cite{kong1992}
of the attraction weights $\alpha$ (Eq.~\eqref{eq:hsw}), a standard
diagnostic in importance sampling that converts a set of weights
into an effective number of equally weighted samples.
For weights normalized to sum to one,
$\mathrm{ESS}(u)=
1/\sum_{v\in\mathcal{S}^{\dagger}(u)}\alpha(v)^2$
equals $1$ when a single sample carries all the weight and
$|\mathcal{S}^{\dagger}(u)|$ when the weights are uniform.
Unlike the set size, the ESS accounts for how unevenly the
weights are distributed. Fig.~\ref{fig:iss_ess} shows its
distribution over query locations sampled from the training
sets~\cite{lsdir,div2k,flickr2k}. Most locations have an ESS between
two and three, indicating that multiple attraction samples
contribute to the drifting field $V$ (Eq.~\eqref{eq:field}).
The paired HQ residual typically receives the largest attraction
weight, $\alpha(u)$ (Fig.~\ref{fig:hsw_analysis}(b)).
The upper tail extends toward the maximum of $K+1$, indicating
contributions from more samples. Values near one indicate
concentration on a single sample, including singleton fallback
cases.

\subsubsection{HSW: sensitivity of the drifting field.}
\label{sec:hsw_analysis}

\begin{figure}[h]
    \centering
    \includegraphics[width=1\linewidth]{\OtterArchiveRoot 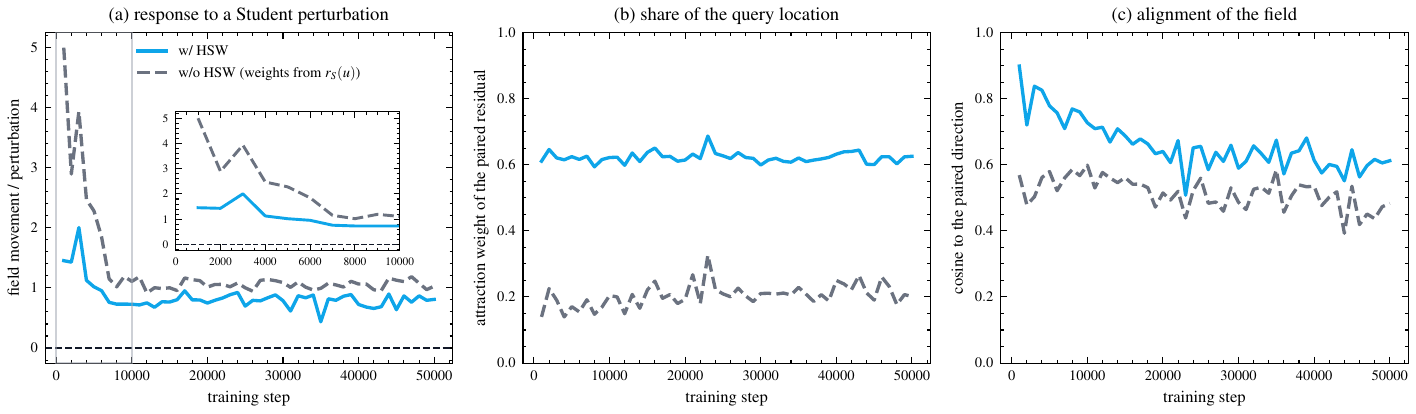}
    \caption{\textbf{Sensitivity of the drifting field
$V$ (Eq.~\eqref{eq:field}).} At evaluation points during training,
we add small Gaussian noise to the DAR latent and recompute the
drifting field under each weighting rule using the same DAR state.
The perturbations are applied only for evaluation.
(a) The change in the drifting field relative to the perturbation,
$\lVert V(r_S+\delta)-V(r_S)\rVert_2/\lVert\delta\rVert_2$,
with a zoom on the early phase; a smaller response indicates
lower sensitivity to the perturbation.
(b) The attraction weight $\alpha(u)$ assigned to the paired HQ
residual (Eq.~\eqref{eq:hsw}).
(c) The cosine similarity between $V$ and the direction toward
the paired HQ residual; higher values indicate stronger alignment.}
    \label{fig:hsw_analysis}
\end{figure}

Variant (e) in Tab.~\ref{tab:ablation} replaces
the attraction weights $\alpha$ of HSW (Sec.~\ref{sec:hsw},
Eq.~\eqref{eq:hsw}) with $\alpha$ computed from the DAR residual
$r_S(u)$. It produces the smallest performance loss among the
ablations, with worse results across all reported metrics.
We examine how this weighting choice affects the drifting field
during training.

To measure this sensitivity, we add small Gaussian
noise to the DAR latent at several evaluation points during
training and recompute the drifting field using each weighting
rule. We use the same DAR latent, noise perturbation, and retained
set $\mathcal{S}^{\dagger}(u)$ for both rules, changing only the
attraction weighting rule. Because the drifted target is
$\tilde{r}=r_S+V$ (Eq.~\eqref{eq:target}), the perturbation itself
also appears in the target through $r_S$. We measure only the
change in the drifting field $V$ (Eq.~\eqref{eq:field}),
as shown in Fig.~\ref{fig:hsw_analysis}(a).
With HSW, the response of the drifting field remains small and
decreases over training. With the DAR-dependent $\alpha$, the
response is several times larger early in training, and the gap
narrows as training proceeds.

Fig.~\ref{fig:hsw_analysis}(b) examines the
attraction weights underlying this difference. With HSW,
$\alpha(u)$, the weight assigned to the paired HQ residual,
remains the largest throughout training. With the DAR-dependent
$\alpha$, its weight drops sharply as the weighting favors
attraction samples $r_H(v)\in\mathcal{A}(u)$ that resemble the
current DAR residual (Sec.~\ref{sec:hsw}).
Fig.~\ref{fig:hsw_analysis}(c) compares the resulting field
directions. With HSW, the drifting field $V$ is initially aligned
with the direction toward the paired HQ residual and departs from
it smoothly as training proceeds. Without HSW, this early
alignment is weaker. Together, these results show that HSW
reduces the field's sensitivity to DAR perturbations and maintains
stronger alignment with the direction toward the paired HQ
residual early in training.

\subsubsection{HPR: the regression space.}
\label{sec:hpr_analysis}

Variant (a) in Tab.~\ref{tab:ablation} replaces
HPR (Sec.~\ref{sec:hpr}) with latent-space regression.
It produces the largest performance loss among the ablations,
with lower fidelity and perceptual quality. HPR separates
the space in which the drifting field is constructed from
the space in which the regression is measured.
The drifting field $V$ (Eq.~\eqref{eq:field}) remains in the RRS.
Regression toward the drifted target $\tilde{r}$
(Eq.~\eqref{eq:target}) is performed after decoding in the
pixel space (Eq.~\eqref{eq:hpr_decode}).

To isolate the effect of the regression space alone,
we compare two controlled runs without the HF split used in
the full method (Fig.~\ref{fig:latent_vs_pixel}).
The two runs use the same seed and configuration and construct
the drifted target in the same way. The only difference is
whether regression is measured in the latent space or the
pixel space. Latent-space regression produces grid-like
artifacts, whereas pixel-space regression restores texture
without these artifacts. Together with Variant (a), this
comparison supports measuring regression on the decoded images
while retaining the drifting field in RRS.
The full method further restricts this pixel-space regression
to the HF band through $\mathcal{L}_{\mathrm{HPR}}$
(Eq.~\eqref{eq:hpr_loss}) to focus the supervision on fine
details, while paired reconstruction losses supervise
overall fidelity to the HQ image.

\section{Conclusion}
\label{sec:conclusion}

We have presented \ourmethod{}, a teacher-free framework for
drifting-based one-step paired image restoration.
NSD addresses \textit{conditional sample collapse}, where each
LQ input has only one paired HQ target and one deterministic
DAR output, by constructing internal surrogate samples.
RRS and ISS form attraction and repulsion sample sets in
residual space, while HSW computes HQ-based attraction weights
and HPR performs regression toward the drifted target on the
HF band of decoded images.
The resulting objective provides distribution-level supervision
without a Teacher Model, an Auxiliary Network, a Discriminator,
or alternating optimization.

Experiments on Real-ISR and all-in-one restoration, together
with component ablations, have demonstrated the effectiveness
of \ourmethod{}.
Its application to U-Net, DiT, and an unmodified Restormer
trained from scratch further supports its applicability
across backbones.
NSD relies on internal recurrence: when no additional candidate
passes validity filtering, the drifted residual target reduces
to the paired HQ residual.
Extending sample construction beyond a single image and
applying NSD to video restoration are promising directions
for future work.

\section*{Acknowledgments}
This work was supported by the Ministry of Education of the Republic of Korea and the National Research Foundation of Korea (NRF-2025S1A5C3A04022639). This work was partly supported by Institute of Information \& communications Technology Planning \& Evaluation (IITP) grant funded by the Korea government(MSIT) (No. RS-2025-25422680, Metacognitive AGI Framework and its Applications, 50\%).

\pdfbookmark[1]{References}{otter-references}
{\small

}
\clearpage
\pdfbookmark[1]{Supplementary Material}{otter-supplement}
\twocolumn[{\centering\Large\bfseries Supplementary Material\par\vspace{1em}}]
\providecommand{\ottercheck}[1]{\par\noindent
  \textcolor{red}{\textbf{Author check.} #1}\par}
\providecommand{\otterTBD}{\textcolor{red}{TBD}}
\providecommand{\ottervariant}[1]{OTTER$_{\mathrm{B}(\text{#1})}$}
\providecommand{\otterfigslot}[1]{%
  \fbox{\parbox{0.94\linewidth}{\centering\small
    \textcolor{red}{Figure pending: #1}}}}

\appendix
\setcounter{equation}{0}
\renewcommand{\theequation}{S\arabic{equation}}
\setcounter{table}{0}
\renewcommand{\thetable}{S\arabic{table}}
\setcounter{figure}{0}
\renewcommand{\thefigure}{S\arabic{figure}}
% Separate PDF hyperlink destinations after resetting counters.
\ifdefined\theHequation
  \renewcommand{\theHequation}{supp.\arabic{equation}}
\fi
\ifdefined\theHtable
  \renewcommand{\theHtable}{supp.\arabic{table}}
\fi
\ifdefined\theHfigure
  \renewcommand{\theHfigure}{supp.\arabic{figure}}
\fi

% =========================================================
\section{Preliminaries and Notation}
\label{supp:prelim}

\subsection{Diffusion Models and Flow Matching}
\label{supp:prelim_dm}

Let $z_0$ denote a clean latent and
$\eta \sim \mathcal{N}(0,I)$ denote Gaussian noise.
A signal-noise path takes the form
\begin{equation}
    z_t = a_t z_0 + b_t \eta,
    \label{eq:supp_path}
\end{equation}
where $a_t$ and $b_t$ specify the signal and noise schedules.
A noise-prediction diffusion model learns to predict the
noise $\eta$ in Eq.~\eqref{eq:supp_path} from $(z_t,t)$,
with conditioning inputs suppressed in the
notation.
We use $a_t$ and $b_t$ to distinguish these schedules from
the attraction and repulsion weights $\alpha$ and $\beta$
used in NSD.

For rectified flow, we adopt a data-to-noise time convention
in the following formulation: $t=0$ corresponds to the clean
latent and $t=1$ to noise.
The linear interpolation and its pathwise velocity are
\begin{equation}
    z_t = (1-t)z_0 + t\eta,
    \qquad
    \frac{\mathrm{d}z_t}{\mathrm{d}t} = \eta-z_0.
    \label{eq:supp_flow}
\end{equation}
A velocity predictor learns the pathwise target
$\eta-z_0$ in Eq.~\eqref{eq:supp_flow}, and generation
integrates the learned field from noise toward
data.

These formulations describe the generative pre-training
of the backbone.
OTTER adapts the pre-trained backbone to paired image
restoration using NSD and paired reconstruction losses.
The resulting one-step restoration mapping is described
in Sec.~\ref{supp:prelim_onestep}.

\subsection{One-Step Distillation Setting}
\label{supp:prelim_onestep}

Given a paired input $(x_L,x_H)$, the frozen VAE encoder
produces $z_L=\mathcal{E}(x_L)$ and $z_H=\mathcal{E}(x_H)$.
The Drift-Aligned Restorer (DAR) performs the one-step mapping
\begin{equation}
    z_S = G_\theta(z_L),
    \qquad
    x_S = \mathcal{D}(z_S).
    \label{eq:supp_onestep}
\end{equation}
In Eq.~\eqref{eq:supp_onestep}, $G_\theta$ includes the
backbone-specific conversion from the native network
prediction to the restored latent.

We adapt the pre-trained U-Net and DiT backbones using
low-rank adaptation (LoRA).
For these diffusion variants, $\theta$ denotes only the
trainable LoRA parameters.
The original backbone weights and both VAE components
remain frozen throughout training.
The decoder remains differentiable with respect to $z_S$,
allowing image-space losses to update $\theta$ through
the decoder.

During training, NSD constructs the drifted target from
the paired HQ image and the current DAR output.
We detach the target branch and update the DAR using NSD
and paired reconstruction losses.
At inference, restoration uses the LQ input alone and
requires one restoration-backbone evaluation together
with VAE encoding and decoding.
ISS retrieval and drifted-target construction operate
only during training.

Backbone-specific implementation details are provided
in Sec.~\ref{supp:impl}.
The conventional Restormer setting is described separately
in Sec.~\ref{supp:impl_cnn}.

\subsection{Notation Table}
\label{supp:prelim_notation}

Tab.~\ref{tab:supp_notation} summarizes the notation used
in the main paper and supplementary material.
As in the main paper, the dependence of $\alpha(v)$ and
$\beta(v)$ on the query location $u$ is left implicit.

\begin{table*}[h]
\caption{Summary of notation. Query and candidate locations
belong to the same training image. $K$ excludes the query
location, so $|\mathcal{S}(u)|=K+1$ before validity filtering.}
\label{tab:supp_notation}
\centering
\small
\begin{tabular}{p{0.23\textwidth}p{0.70\textwidth}}
\toprule
Symbol & Description \\
\midrule
$x_L,x_H,x_S$
& LQ image, paired HQ image, and restored image \\

$\mathcal{E},\mathcal{D}$
& Frozen VAE encoder and decoder in the diffusion setting \\

$G_\theta,\theta$
& DAR and its trainable parameters \\

$z_L,z_H,z_S$
& LQ, HQ, and DAR latents \\

$r_H,r_S$
& HQ and DAR residuals,
$r_H=z_H-z_L$ and $r_S=z_S-z_L$ \\

$\Omega$
& Spatial locations on the two-dimensional latent grid \\

$u,v$
& Query and candidate spatial locations \\

$\mathcal{W}(u)$
& Local search window centered at $u$ \\

$p_H(u)$
& Vectorized $3\times3$ patch centered at $u$ in the HQ residual map \\

$d(u,v)$
& HQ residual patch distance $\|p_H(u)-p_H(v)\|_2$, used for
retrieval, validity filtering, and attraction weighting \\

$K$
& Number of nearest non-query candidates retrieved
before validity filtering \\

$\mathcal{S}(u)$
& Retrieved locations including $u$,
with $|\mathcal{S}(u)|=K+1$ \\

$\mathcal{S}^{\dagger}(u)$
& Locations retained after validity filtering;
$u$ is always retained \\

$\gamma_{\mathrm{rel}},\gamma_{\mathrm{abs}}$
& Relative and absolute validity thresholds in ISS \\

$\mathcal{A}(u)$
& HQ attraction samples,
$\{r_H(v)\}_{v\in\mathcal{S}^{\dagger}(u)}$ \\

$\mathcal{R}(u)$
& DAR repulsion samples,
$\{r_S(v)\}_{v\in\mathcal{S}^{\dagger}(u)}$;
the repulsion sum excludes $v=u$ \\

$\alpha(v),\beta(v)$
& Attraction and repulsion weights for query $u$,
respectively \\

$\tau$
& Attraction-kernel bandwidth in HSW \\

$V(u)$
& Drifting field at query location $u$ \\

$\tilde r(u),\tilde r$
& Drifted residual target $\tilde r(u)=r_S(u)+V(u)$
and the assembled target map \\

$\tilde x$
& Decoded drifted target,
$\tilde x=\mathcal{D}(z_L+\tilde r)$ \\

$b_\sigma,\sigma$
& Gaussian blur operator and its standard deviation \\

$\mathrm{HF}$
& High-frequency extraction,
$\mathrm{HF}(x)=x-b_\sigma(x)$ \\

$\mathcal{L}_{\mathrm{HPR}}$
& High-frequency pixel-space regression loss \\

$\mathcal{L}_{\mathrm{LPIPS}}$
& Paired LPIPS reconstruction loss \\

$\mathcal{L}$
& Total objective combining HPR and paired
reconstruction losses \\

$\lambda_1,\lambda_2$
& Weights of the paired $\ell_1$ and LPIPS losses,
respectively \\

$\mathrm{ESS}(u)$
& Effective sample size of the normalized attraction
weights at $u$ \\

$\delta$
& Gaussian perturbation of the DAR latent used
in the sensitivity analysis \\

$\operatorname{sg}[\cdot]$
& Stop-gradient operator \\

\midrule
$z_0,z_t$
& Clean latent and intermediate latent
on the signal-noise path \\

$\eta,t$
& Gaussian noise and generative time variable \\

$a_t,b_t$
& Signal and noise schedules
in Eq.~\eqref{eq:supp_path} \\
\bottomrule
\end{tabular}
\end{table*}

\section{Normalization of the Attraction and Repulsion Weights}
\label{supp:weights}

As in the main paper, we write $\alpha(v)$ and $\beta(v)$
with the query index $u$ left implicit.
Both weights are normalized over the retained locations
$\mathcal{S}^{\dagger}(u)$, including the query location $u$.

HSW uses the HQ residual patch distance $d(u,v)$
defined in Eq.~\eqref{eq:iss_dist} of the main paper.
For each $v\in\mathcal{S}^{\dagger}(u)$, we normalize the
attraction kernel in Eq.~(8) of the main paper as
\begin{equation}
    \alpha(v)=
    \frac{\exp\!\left(-d(u,v)/\tau\right)}
    {\displaystyle\sum_{w\in\mathcal{S}^{\dagger}(u)}
     \exp\!\left(-d(u,w)/\tau\right)}.
    \label{eq:supp_attraction_weights}
\end{equation}
Equation~\eqref{eq:supp_attraction_weights} gives
$\sum_{v\in\mathcal{S}^{\dagger}(u)}\alpha(v)=1$.
We use these same weights in the drifting field
(Eq.~(7) of the main paper) and the ESS analysis
(Sec.~4.3.2 of the main paper).
Since $d(u,u)=0$, the paired HQ residual receives a maximum
attraction weight.
For a fixed training pair, both the retained locations and
the attraction weights remain independent of the DAR output.

For repulsion, we use distances between the DAR
residual vectors at the center locations with the same
bandwidth $\tau$:
\begin{equation}
    \beta(v)=
    \frac{\exp\!\left(-\|r_S(u)-r_S(v)\|_2/\tau\right)}
    {\displaystyle\sum_{w\in\mathcal{S}^{\dagger}(u)}
     \exp\!\left(-\|r_S(u)-r_S(w)\|_2/\tau\right)}.
    \label{eq:supp_repulsion_weights}
\end{equation}
Equation~\eqref{eq:supp_repulsion_weights} normalizes the
DAR-dependent weights over the same retained locations,
giving $\sum_{v\in\mathcal{S}^{\dagger}(u)}\beta(v)=1$.
The repulsion sum in Eq.~(7) of the main paper omits $v=u$
because its displacement $r_S(u)-r_S(u)$ is zero.
The remaining weights retain the normalization in
Eq.~\eqref{eq:supp_repulsion_weights} and therefore sum to
$1-\beta(u)$; they are not renormalized after excluding
the self-displacement.

\subsection{Singleton Fallback}
\label{supp:weights_singleton}

ISS always retains the query location.
Consequently, the denominators in
Eqs.~\eqref{eq:supp_attraction_weights}
and~\eqref{eq:supp_repulsion_weights} each contain a
self-kernel value of one and remain well-defined.

When no additional candidate passes validity filtering,
$\mathcal{S}^{\dagger}(u)=\{u\}$ and both normalizations give
$\alpha(u)=\beta(u)=1$.
The repulsion sum is empty, so Eq.~(7) of the main paper
and the drifted-target definition give
\begin{equation}
    V(u)=r_H(u)-r_S(u),
    \qquad
    \tilde r(u)=r_H(u).
    \label{eq:supp_singleton}
\end{equation}
Equation~\eqref{eq:supp_singleton} shows that the drifted
residual target at a singleton location reduces to the
paired HQ residual.
If every query location is a singleton, the assembled
target satisfies $\tilde r=r_H$ and decodes to
$\mathcal{D}(z_H)$.
HPR then supervises the high-frequency components of
this decoded HQ latent.
The paired reconstruction losses continue to compare
$x_S$ with the original HQ image $x_H$, which need not
equal $\mathcal{D}(z_H)$ exactly.

\subsection{Consistency of the Drifting Field and HPR}
\label{supp:weights_zero}

Suppose $r_S(v)=r_H(v)$ for every retained location
$v\in\mathcal{S}^{\dagger}(u)$, including the query $u$.
The displacement at $u$ vanishes, and the field in
Eq.~\eqref{eq:field} reduces to
\begin{equation}
    V(u)=\sum_{v\in\mathcal{S}^{\dagger}(u)\setminus\{u\}}
    (\alpha(v)-\beta(v))\bigl(r_H(v)-r_H(u)\bigr).
    \label{eq:supp_matched_residuals}
\end{equation}
Equation~\eqref{eq:supp_matched_residuals} gives zero drift
when the normalized weights also match. Equality of center
residuals alone does not imply equal weights because the
attraction kernel uses patch descriptors, while the
repulsion kernel uses center residual vectors.

When $V(u)=0$ for every $u\in\Omega$, the drifted target
satisfies $\tilde r=r_S$.
Decoding both branches with the same frozen decoder gives
\begin{equation}
    z_L+\tilde r=z_S,
    \qquad
    \tilde x=x_S,
    \qquad
    \mathcal{L}_{\mathrm{HPR}}=0.
    \label{eq:supp_zero_field}
\end{equation}
Equation~\eqref{eq:supp_zero_field} establishes that a
zero drifting field yields zero HPR loss.
Its converse is not guaranteed because HPR compares only
the high-frequency components after decoding.
This property concerns HPR; the paired reconstruction
losses remain separately defined with respect to $x_H$.

\section{Implementation Details}
\label{supp:impl}

\subsection{Training Algorithm}
\label{supp:impl_algo}

Algorithm~\ref{alg:supp_training} summarizes OTTER training
for the diffusion variants.
As described in Sec.~\ref{supp:prelim_onestep}, we update
only the LoRA parameters $\theta$, while keeping the original
backbone weights and both VAE components frozen.
ISS retrieves candidates within each training image.
The attraction and repulsion weights follow
Sec.~\ref{supp:weights}.

We detach the drifted-target branch using the stop-gradient
operator $\operatorname{sg}[\cdot]$.
The prediction branch retains gradients through the frozen
decoder, allowing the image-space losses to update the DAR.
The total objective follows Eq.~(12) of the main paper.

\begin{algorithm}[h]
\caption{OTTER training for the diffusion variants.}
\label{alg:supp_training}
\begin{algorithmic}[1]
\Require Paired dataset $\{(x_L,x_H)\}$;
frozen VAE $(\mathcal{E},\mathcal{D})$;
DAR $G_\theta$ with trainable LoRA parameters $\theta$;
NSD hyperparameters and loss weights

\While{training}
    \State Sample a minibatch of paired images
    \State $z_L\gets\mathcal{E}(x_L)$,
           $z_H\gets\mathcal{E}(x_H)$
    \State $z_S\gets G_\theta(z_L)$,
           $x_S\gets\mathcal{D}(z_S)$
    \State $r_H\gets z_H-z_L$,
           $r_S\gets z_S-z_L$

    \For{each image and each $u\in\Omega$ in parallel}
        \State Retrieve $K$ non-query neighbors using
               $d(u,v)$ and include $u$ to form
               $\mathcal{S}(u)$
        \State Apply validity filtering to obtain
               $\mathcal{S}^{\dagger}(u)$
        \State Form $\mathcal{A}(u)$ and $\mathcal{R}(u)$
               from the retained locations
        \State Compute $\alpha(v)$ and $\beta(v)$ using
               Eqs.~\eqref{eq:supp_attraction_weights}
               and~\eqref{eq:supp_repulsion_weights}
        \State Compute $V(u)$ using Eq.~(7) of the main paper
        \State $\tilde r(u)\gets r_S(u)+V(u)$
    \EndFor

    \State Assemble the drifted target map
           $\tilde r=[\tilde r(u)]_{u\in\Omega}$
    \State $\tilde x\gets
           \operatorname{sg}[\mathcal{D}(z_L+\tilde r)]$
    \State $\mathcal{L}_{\mathrm{HPR}}\gets
           \|\mathrm{HF}(x_S)-\mathrm{HF}(\tilde x)\|_1$
    \State $\mathcal{L}\gets
           \mathcal{L}_{\mathrm{HPR}}
           +\lambda_1\|x_S-x_H\|_1$
    \Statex \hspace{\algorithmicindent}
           $\phantom{\mathcal{L}\gets}\quad
           +\lambda_2\mathcal{L}_{\mathrm{LPIPS}}(x_S,x_H)$
    \State Update $\theta$ using $\nabla_\theta\mathcal{L}$
\EndWhile
\end{algorithmic}
\end{algorithm}

\subsection{Training Configuration for Real-ISR}
\label{supp:impl_realisr}

We train the Real-ISR variants on LSDIR, DIV2K, and Flickr2K.
LQ-HQ training pairs are synthesized using the
Real-ESRGAN degradation pipeline.
The U-Net and DiT variants use the same pre-trained
backbone models as PiSA-SR and TSD-SR, respectively.
We optimize only the LoRA parameters, while keeping
the original backbone weights and both VAE components frozen.

Tab.~\ref{tab:supp_realisr_training} summarizes the
training configurations for the two diffusion backbones.
The hyperparameters of NSD are provided separately
in Sec.~\ref{supp:impl_hyper}.

\begin{table}[h]
\caption{Training configurations for Real-ISR.
The U-Net and DiT settings follow PiSA-SR~\cite{pisasr}
and TSD-SR~\cite{tsdsr}, respectively.}
\label{tab:supp_realisr_training}
\centering
\small
\begin{tabular}{lcc}
\toprule
Setting & U-Net & DiT \\
\midrule
Optimizer          & AdamW & AdamW \\
Learning rate      & $5\times10^{-5}$ & $5\times10^{-6}$ \\
Batch size (total) & 16 & 16 \\
HQ crop size       & $512\times512$ & $512\times512$ \\
LoRA rank          & 4 & 64 \\
\bottomrule
\end{tabular}
\end{table}

\subsection{Training Configuration for All-in-One Restoration}
\label{supp:impl_cdd}

We train one all-in-one restoration model per backbone on
CDD-11, which contains eleven settings involving four
degradation factors and their combinations.
Training pairs from all eleven settings are combined into
a single pool, from which minibatches are sampled at random.

The U-Net and DiT variants use the backbone initialization
and LoRA adaptation described in
Sec.~\ref{supp:impl_realisr}.
Only the LoRA parameters are updated, while the original
backbone weights and both VAE components remain frozen.
We also train an unmodified Restormer from scratch on
the same collection of degradation settings.
Its integration with NSD is detailed in
Sec.~\ref{supp:impl_cnn}.

Tab.~\ref{tab:supp_train_config} summarizes the CDD-11 training
configurations. Tab.~\ref{tab:supp_hyper} lists the NSD
hyperparameters shared across backbones.
\begin{table}[h]
\caption{Training configurations for all-in-one restoration
on CDD-11. N/A denotes not applicable.}
\label{tab:supp_train_config}
\centering
\small
\begin{tabular}{lccc}
\toprule
Setting & U-Net & DiT & Restormer \\
\midrule
Optimizer          & AdamW & AdamW & AdamW \\
Learning rate      & $5\times10^{-5}$ & $5\times10^{-6}$ & $5\times10^{-5}$ \\
Batch size (total) & 16 & 16 & 16 \\
LoRA rank          & 4 & 64 & N/A \\
\bottomrule
\end{tabular}
\end{table}

\subsection{Hyperparameters of NSD}
\label{supp:impl_hyper}

Tab.~\ref{tab:supp_hyper} summarizes the NSD hyperparameters.
ISS uses a $15\times15$ search window on the latent grid
and retrieves $K=10$ non-query locations.
Including the query itself yields
$|\mathcal{S}(u)|=K+1=11$ candidates before validity filtering.
Filtering removes invalid non-query candidates while
always retaining the query.

\begin{table}[h]
\caption{NSD hyperparameters.}
\label{tab:supp_hyper}
\centering
\footnotesize
\begin{tabular}{p{0.56\columnwidth}lc}
\toprule
Hyperparameter & Symbol & Value \\
\midrule
ISS window side length (latent cells) & --- & 15 \\
ISS descriptor patch size & --- & $3\times3$ \\
Non-query retrieved candidates & $K$ & 10 \\
Relative validity threshold & $\gamma_{\mathrm{rel}}$ & 2.0 \\
Absolute validity threshold & $\gamma_{\mathrm{abs}}$ & 1.2 \\
HSW attraction bandwidth & $\tau$ & 0.2 \\
HPR Gaussian standard deviation & $\sigma$ & 1.5 \\
\bottomrule
\end{tabular}
\end{table}

\subsection{Training Overhead of NSD}
\label{supp:impl_overhead}

In the diffusion setting, \ourmethod{} requires one
diffusion-scale network forward pass per training iteration.
Other training computations include VAE encoding, ISS retrieval,
drifting-field estimation, and paired perceptual-loss evaluation.
Both the DAR prediction and the drifted target are decoded
using the frozen VAE decoder.
The target branch is computed without gradient tracking,
while gradients propagate through the decoder on the
prediction branch.
ISS and drifted-target construction are used only during
training and introduce no additional inference cost.

\subsection{Details of the Conventional Restorer}
\label{supp:impl_cnn}
\label{supp:impl_restormer}

We train an unmodified Restormer from scratch on CDD-11,
optimizing all network parameters.
The network directly predicts a restored RGB image.
During training, the frozen VAE encodes the LQ image,
the paired HQ image, and a detached, range-clipped copy
of the Restormer output.
The resulting latents provide the residual representations
for sample retrieval and drifting-field estimation.

HPR and paired reconstruction losses are evaluated directly
on the Restormer output.
All VAE operations and target construction are performed
without gradient tracking, so the losses update Restormer
without backpropagating through the VAE.
At inference, only the Restormer network is required.

\section{Hyperparameter Sensitivity}
\label{supp:sensitivity}

We evaluate the sensitivity of the U-Net variant to $K$
and $\tau$ on RealSR.
We vary $K$ with $\tau=0.2$ fixed and vary $\tau$ with
$K=10$ fixed.
Tab.~\ref{tab:supp_sensitivity} shows modest performance
variation across $K\in\{5,10,15\}$ and
$\tau\in\{0.1,0.2,0.5,1.0\}$.
The PSNR range remains below $0.3$\,dB in each sweep,
while LPIPS and pFID also remain comparable.
These results indicate limited sensitivity to the candidate
count and kernel bandwidth within the evaluated ranges.

\begin{table}[h]
\caption{Hyperparameter sensitivity of the U-Net variant on
RealSR. The first row is the default configuration reported
in Tab.~\ref{tab:main}.}
\label{tab:supp_sensitivity}
\centering
\footnotesize
\begin{tabular}{ccccc}
\toprule
$K$ & $\tau$ & PSNR$\uparrow$ & LPIPS$\downarrow$ & pFID$\downarrow$ \\
\midrule
10 & 0.2 & 25.93 & 0.2576 & 99.58 \\
\midrule
5  & 0.2 & 25.72 & 0.2574 & 99.29 \\
15 & 0.2 & 25.76 & 0.2567 & 99.36 \\
\midrule
10 & 0.1 & 25.64 & 0.2539 & 98.42 \\
10 & 0.5 & 25.85 & 0.2608 & 100.18 \\
10 & 1.0 & 25.79 & 0.2554 & 98.94 \\
\bottomrule
\end{tabular}
\end{table}

\section{Additional Qualitative Results}
\label{supp:qual}

\begin{figure*}[h]
\centering
\includegraphics[width=\textwidth]{\OtterArchiveRoot 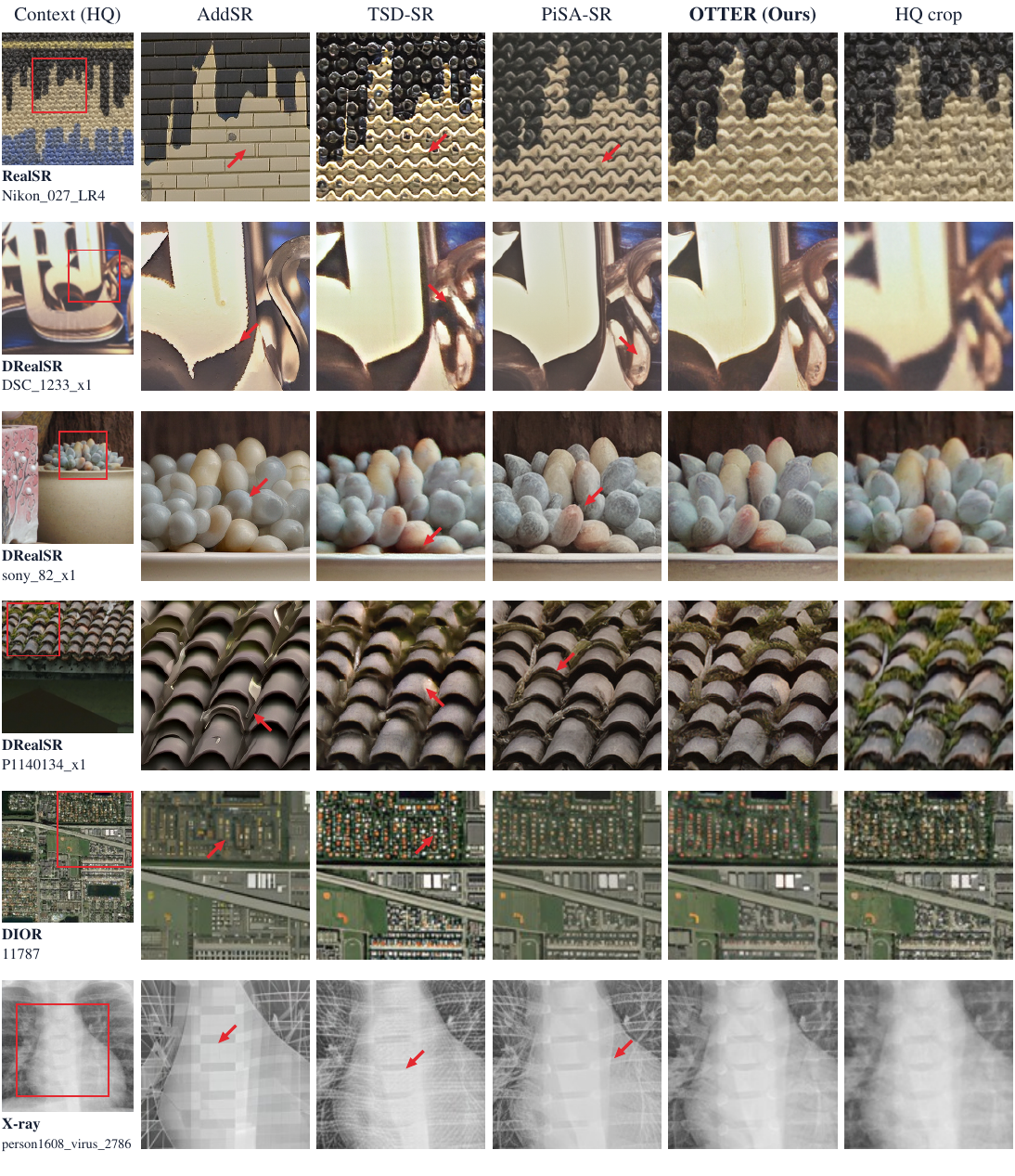}
\caption{Additional super-resolution comparisons using our
U-Net variant on RealSR, DRealSR, DIOR, and chest X-ray imagery.
The left column shows the HQ image for spatial context,
with red boxes marking the enlarged regions.
Red arrows highlight representative artifacts and structural
discrepancies in competing results.
The rightmost column shows the corresponding HQ crops.}
\label{fig:supp_realisr}
\end{figure*}

\begin{figure*}[h]
\centering
\includegraphics[width=\textwidth]{\OtterArchiveRoot 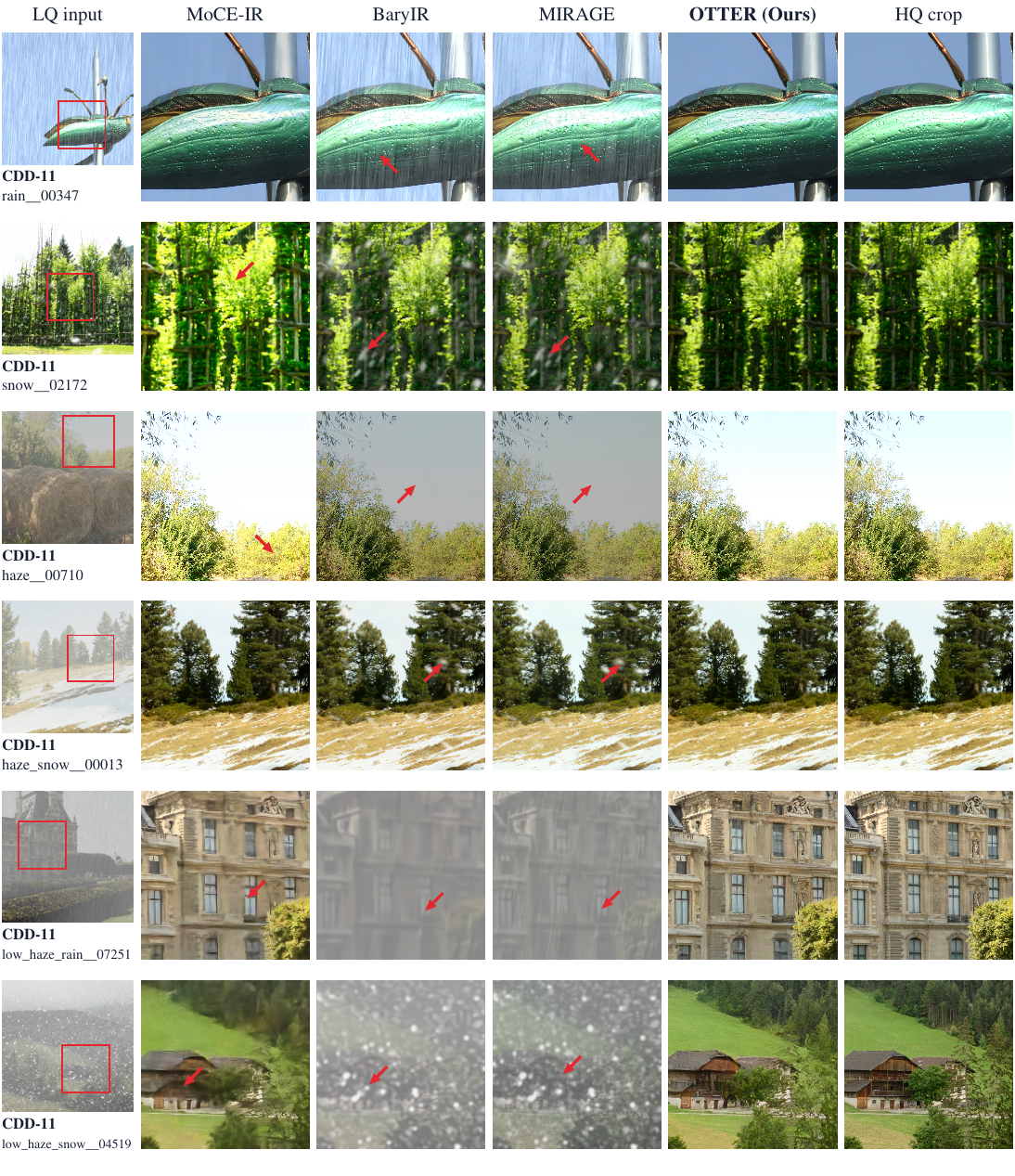}
\caption{Additional all-in-one restoration comparisons on
CDD-11 using our Restormer variant.
Rows show rain, snow, haze, haze--snow,
low-light--haze--rain, and low-light--haze--snow settings.
Red boxes in the LQ inputs mark the enlarged regions;
red arrows highlight residual degradations and discrepancies
from the HQ references in competing results.}
\label{fig:supp_cdd}
\end{figure*}

We provide additional qualitative comparisons for Real-ISR
and all-in-one restoration in
Figs.~\ref{fig:supp_realisr} and~\ref{fig:supp_cdd}, respectively.
Red arrows highlight representative artifacts, residual
degradations, and discrepancies from the HQ references.

\paragraph{Real-ISR.}
Fig.~\ref{fig:supp_realisr} compares our U-Net variant with
AddSR, TSD-SR, and PiSA-SR on RealSR and DRealSR, together
with examples from additional image domains.
The knitted pattern illustrates how visually sharp outputs
can contain structures inconsistent with the reference:
competing methods replace the stitches with geometric patterns
or exaggerate their contours.
In comparison, \ourmethod{} more closely preserves the
appearance and organization of the recurring stitches.
The succulent and roof examples further reveal differences
in surface texture and local structure.
Some competing outputs either smooth away texture or introduce
excessive high-frequency details, including granular surfaces
and unnaturally emphasized boundaries.
Our results retain fine details with fewer such distortions,
yielding a closer visual match to the HQ references.

The last two rows show optical remote sensing imagery from
DIOR and a chest X-ray image.
The DIOR example contains densely arranged small structures,
where excessive sharpening introduces artificial repetition
and exaggerated local contrast.
In the X-ray example, several competing outputs exhibit
pronounced banding or grid-like patterns that are absent
from the HQ reference.
\ourmethod{} reduces these artifacts while retaining more
of the reference's spatial organization and smooth intensity
transitions, respectively.
These examples illustrate the importance of controlling
spurious high-frequency content across distinct image domains.

\paragraph{All-in-one restoration.}
Fig.~\ref{fig:supp_cdd} compares our Restormer variant with
MoCE-IR, BaryIR, and MIRAGE on CDD-11.
The examples cover single degradations and combinations
of low light, haze, rain, and snow.
BaryIR and MIRAGE retain visible rain streaks, snow particles,
or a hazy appearance in several examples.
MoCE-IR removes much of the degradation, but some outputs
show exaggerated foliage colors or softened building details.
\ourmethod{} more closely preserves the reference appearance,
including surface reflections, foliage colors, and
architectural structures.
Under triple degradations, it recovers clearer window frames
and roof boundaries while suppressing the degradations
that obscure these structures in the LQ inputs.

\section{User Study}
\label{supp:user_study}

\subsection{Study Protocol}
\label{supp:user_protocol}

We conduct a user study with 33 participants using
20 Real-ISR examples and 20 CDD-11 examples.
The study evaluates our U-Net variant against AddSR,
TSD-SR, and PiSA-SR for Real-ISR, and our Restormer
variant against MoCE-IR, BaryIR, and MIRAGE for
all-in-one restoration.
Each participant evaluates all 40 examples, with
Real-ISR followed by all-in-one restoration.
The recruitment message identified the researcher and their affiliation,
explained the purpose of the study, and invited recipients to
participate voluntarily via a survey link. Participants gave separate
consent to participate. The survey collected ratings of restoration
outputs but no personally identifying information. The study was not
reviewed by an institutional review board (IRB) before data collection.

\begin{figure}[h]
\centering
\includegraphics[width=\columnwidth]{\OtterArchiveRoot 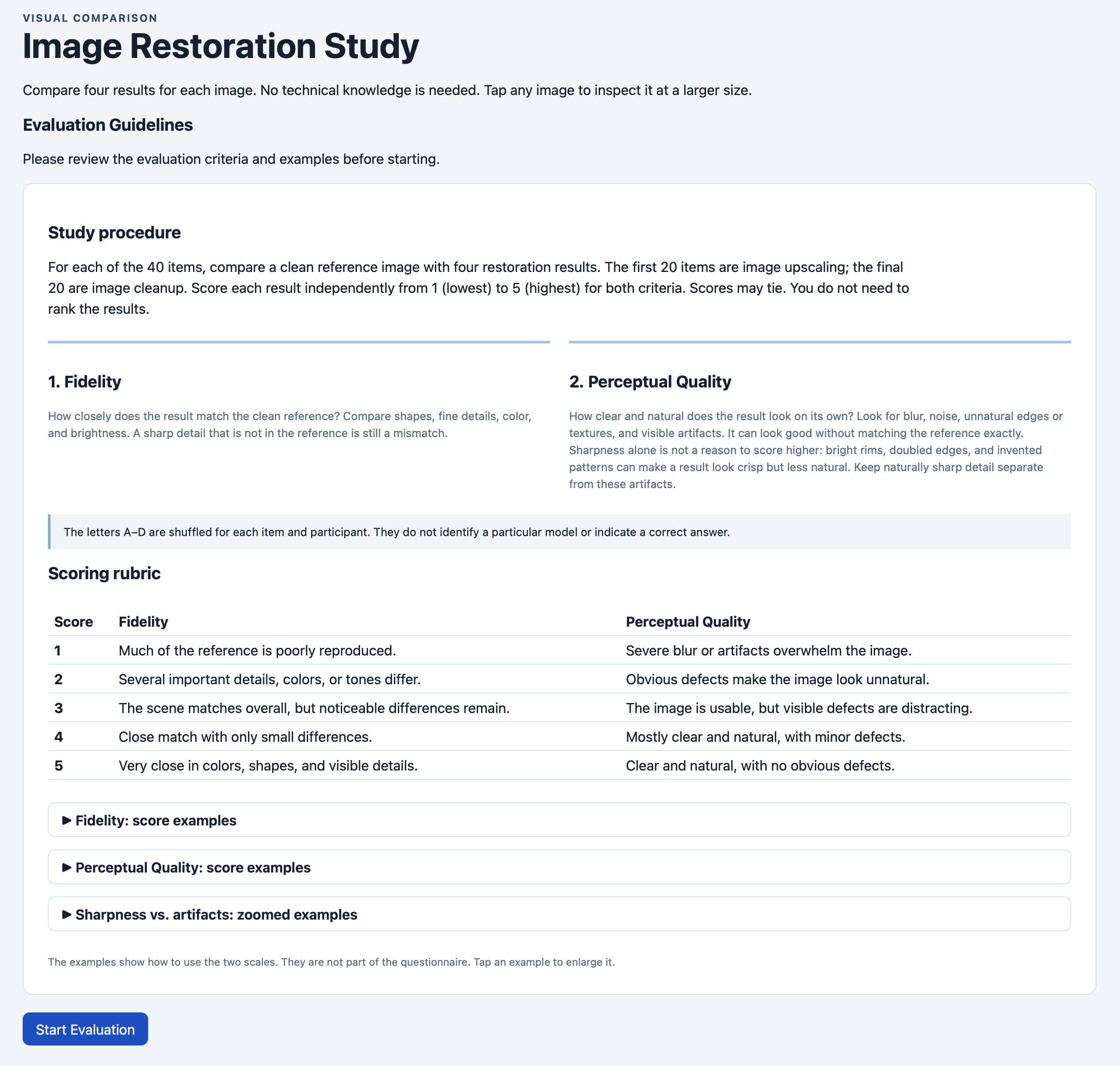}
\caption{User-study guidelines, including the procedure,
evaluation criteria, and five-point scoring rubric.
Expandable examples illustrate the distinction between
sharp details and artifacts.}
\label{fig:supp_user_guide}
\end{figure}

Fig.~\ref{fig:supp_user_guide} shows the evaluation
guidelines presented before the questionnaire.
Participants score each result on two separate
five-point scales.
\emph{Fidelity} measures agreement with the HQ reference
in shape, fine details, color, and brightness.
\emph{Perceptual quality} measures the clarity and
naturalness of the restored image, considering blur,
noise, unnatural textures, and other visible artifacts.
The instructions distinguish naturally sharp details
from artificial contours or invented patterns.
A scoring rubric and illustrative examples explain
both criteria; these examples are separate from
the evaluation items.

\begin{figure}[h]
\centering
\includegraphics[width=\columnwidth]{\OtterArchiveRoot 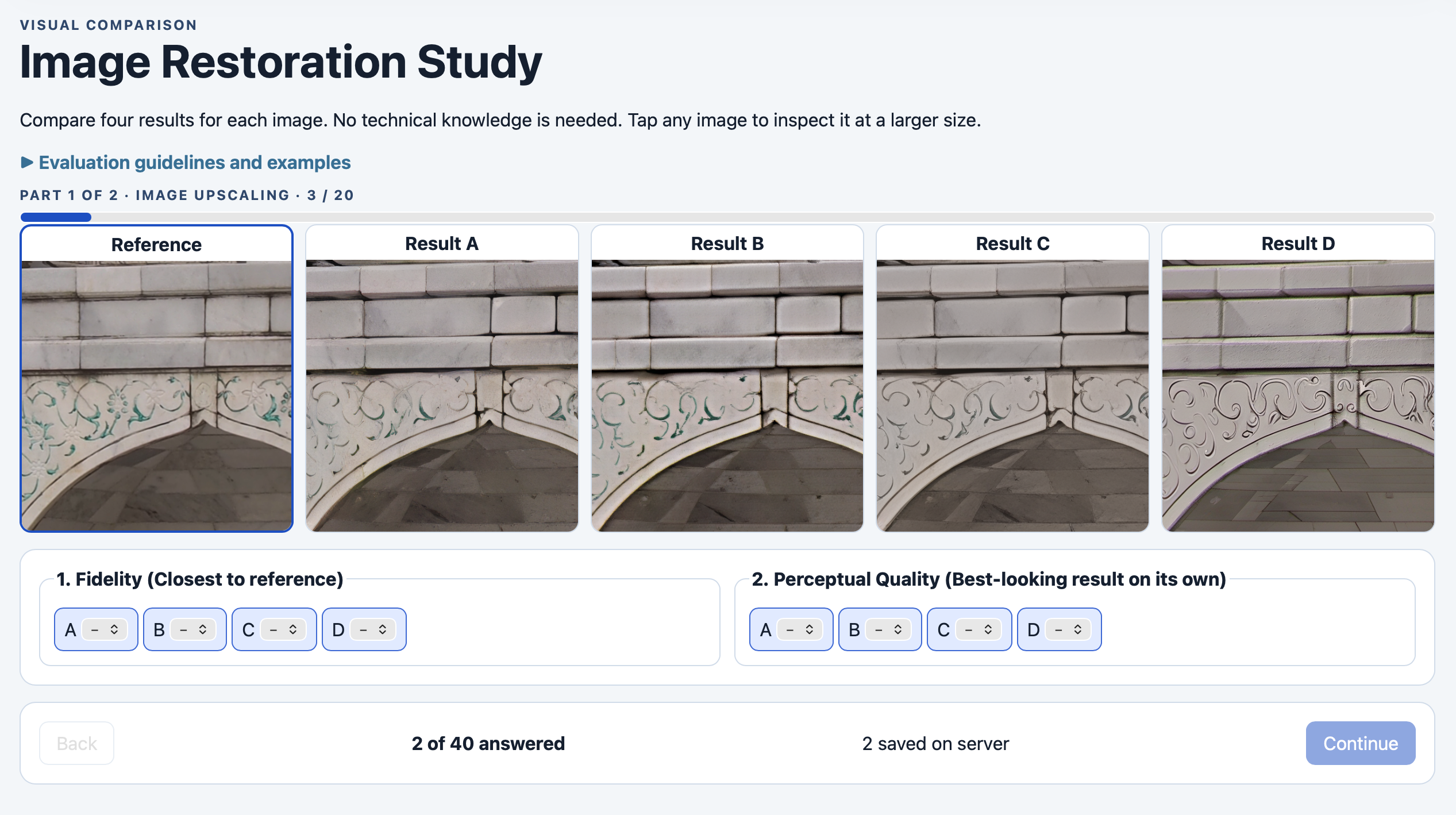}
\caption{User-study interface.
Participants inspect the HQ reference and four
anonymized results, then score fidelity and
perceptual quality separately.
Method assignments to the displayed labels are
shuffled for each participant and example.}
\label{fig:supp_user_interface}
\end{figure}

Fig.~\ref{fig:supp_user_interface} shows the rating
interface.
Each item displays the HQ reference alongside four
restoration results labeled A, B, C, and D.
Model identities are concealed, and the assignment
of methods to these labels is shuffled for each
participant and example.
Participants can enlarge images for closer inspection.
Each result is scored separately for both criteria,
and equal scores are permitted without requiring
a ranking among methods.

\subsection{Rating Distributions and Analysis}
\label{supp:user_analysis}

Each combination of task, method, and criterion receives
$33\times20=660$ ratings, yielding 10,560 individual
scores across the complete study.
The means in Tab.~\ref{tab:user_study} average these
660 ratings.
Since every participant evaluates the same number
of examples, each participant contributes equally
to the reported means.
Tab.~\ref{tab:supp_user_distribution} provides the
complete score distributions, medians, and proportions
of ratings at or above four.
These are descriptive summaries of repeated ratings
from 33 participants.

\paragraph{Real-ISR.}
Our U-Net variant receives mean fidelity and
perceptual-quality scores of 4.04 and 4.06,
respectively, compared with 3.74 and 3.93 for
PiSA-SR, the runner-up on both criteria
(Tab.~\ref{tab:user_study}).
Ratings of four or five account for 75.5\% of
our fidelity scores and 75.6\% of our
perceptual-quality scores, compared with 66.4\%
and 72.1\% for PiSA-SR.
Thus, the difference is more pronounced in fidelity,
while the perceptual-quality distributions are closer.
Although both methods have a median of four,
our fidelity ratings include a larger proportion
of fives: 44.4\% versus 25.8\%.
These observations support improved agreement with
the reference alongside a smaller increase in
perceived visual quality on the evaluated examples.

\paragraph{All-in-one restoration.}
Our Restormer variant receives a mean score of
4.71 for both criteria in Tab.~\ref{tab:user_study}.
Ratings of four or five account for 93.0\%
under each criterion, compared with 63.2\%
for fidelity and 61.7\% for perceptual quality
with MoCE-IR.
The corresponding differences are 29.8 and
31.4 percentage points.
Our median score is five for both criteria,
compared with four for MoCE-IR.
The higher means are therefore accompanied by
a greater concentration of high ratings across
the pooled evaluations.

\paragraph{Connection to the benchmark results.}
The human ratings complement the automatic
evaluations in Tabs.~\ref{tab:main}
and~\ref{tab:cdd11}.
On all three Real-ISR benchmarks, our U-Net
variant achieves lower LPIPS, DISTS, DreamSim,
and pFID than PiSA-SR, consistent in direction
with its higher mean human ratings.
However, PiSA-SR obtains higher MUSIQ scores,
showing that the human-quality ordering is
not reproduced by every automatic metric.
On CDD-11, the higher human ratings of our
Restormer variant accompany improvements over
MoCE-IR in all six full-reference metrics
and pFID across the three degradation groups.
The user study provides complementary evidence
on the selected examples rather than an
image-level correlation analysis between
human ratings and automatic metrics.

\begin{table}[h]
\caption{Complete user-study rating distributions.
Each row contains 660 ratings from 33 participants
evaluating 20 examples.
Columns 1 to 5 report score frequencies.
Med.: median; $\geq4$: percentage of ratings
assigned four or five; F: fidelity;
PQ: perceptual quality.
Mean ratings appear in Tab.~\ref{tab:user_study}.}
\label{tab:supp_user_distribution}
\centering
\footnotesize
\setlength{\tabcolsep}{2.5pt}
\begin{tabular}{@{}llrrrrrcc@{}}
\toprule
Method & &
\multicolumn{5}{c}{Score counts} &
Med. & $\geq4$ (\%) \\
\cmidrule(lr){3-7}
& & 1 & 2 & 3 & 4 & 5 & & \\
\midrule
\multicolumn{9}{l}{\textit{Real-ISR (our U-Net variant)}} \\
\midrule
\multirow{2}{*}{AddSR}
& F  & 103 & 246 & 240 & 55  & 16  & 2 & 10.8 \\
& PQ & 78  & 116 & 267 & 144 & 55  & 3 & 30.2 \\
\addlinespace
\multirow{2}{*}{TSD-SR}
& F  & 38 & 92 & 176 & 269 & 85 & 4 & 53.6 \\
& PQ & 35 & 95 & 244 & 209 & 77 & 3 & 43.3 \\
\addlinespace
\multirow{2}{*}{PiSA-SR}
& F  & 35 & 48 & 139 & 268 & 170 & 4 & 66.4 \\
& PQ & 33 & 54 & 97  & 216 & 260 & 4 & 72.1 \\
\addlinespace
\multirow{2}{*}{\ourmethod{}}
& F  & 29 & 45 & 88 & 205 & 293 & 4 & \textbf{75.5} \\
& PQ & 26 & 47 & 88 & 200 & 299 & 4 & \textbf{75.6} \\
\midrule
\multicolumn{9}{l}{\textit{CDD-11 (our Restormer variant)}} \\
\midrule
\multirow{2}{*}{BaryIR}
& F  & 239 & 126 & 125 & 105 & 65 & 2 & 25.8 \\
& PQ & 237 & 119 & 135 & 99  & 70 & 2 & 25.6 \\
\addlinespace
\multirow{2}{*}{MIRAGE}
& F  & 226 & 141 & 132 & 103 & 58 & 2 & 24.4 \\
& PQ & 205 & 143 & 153 & 99  & 60 & 2 & 24.1 \\
\addlinespace
\multirow{2}{*}{MoCE-IR}
& F  & 15 & 37 & 191 & 170 & 247 & 4 & 63.2 \\
& PQ & 22 & 60 & 171 & 161 & 246 & 4 & 61.7 \\
\addlinespace
\multirow{2}{*}{\ourmethod{}}
& F  & 5 & 6  & 35 & 82 & 532 & 5 & \textbf{93.0} \\
& PQ & 6 & 10 & 30 & 80 & 534 & 5 & \textbf{93.0} \\
\bottomrule
\end{tabular}
\end{table}

\section{Limitation}
\label{supp:failure}

\begin{figure*}[h]
\centering
\includegraphics[width=\textwidth]{\OtterArchiveRoot 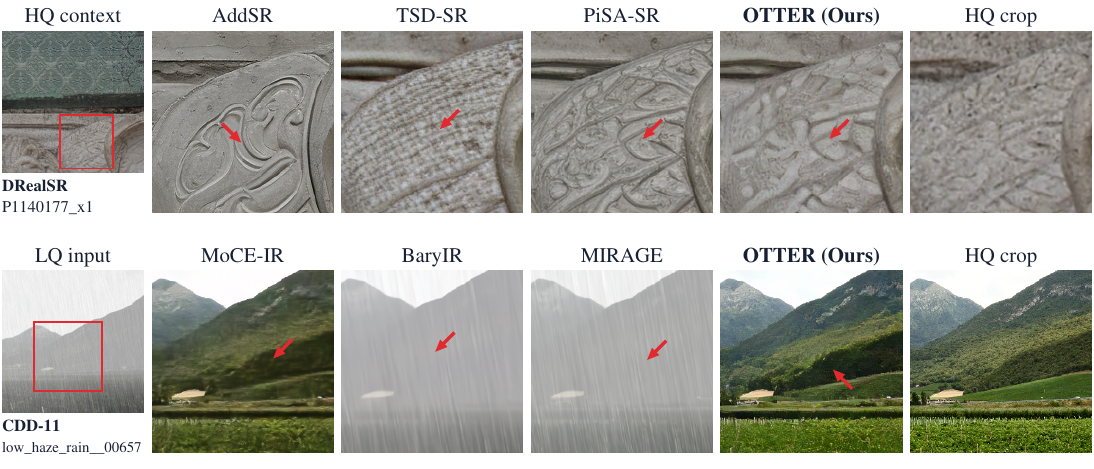}
\caption{Challenging examples on DRealSR (top, our U-Net
variant) and CDD-11 (bottom, our Restormer variant).
All compared methods exhibit discrepancies from the HQ
references.
Red arrows highlight incorrect textures, residual
degradations, or missing details, including remaining
errors in our results.
The left column shows the HQ context for the top row
and the LQ input for the bottom row;
red boxes mark the enlarged regions.}
\label{fig:supp_failure}
\end{figure*}

When degradation removes fine details, distinct HQ images can
produce the same LQ observation. The observation alone cannot
identify the original details, so a generative prior may reconstruct
a plausible texture that differs from the true one. This ambiguity
underlies the perception--distortion tradeoff~\cite{pdtradeoff}
and remains a limitation of generative restoration, including
\ourmethod{}.
Fig.~\ref{fig:supp_failure} presents two examples in which
all compared methods, including \ourmethod{}, fail to fully
reproduce the HQ details.
In the super-resolution example, the methods generate
surface patterns that differ from the irregular texture
of the reference.
Although \ourmethod{} produces less exaggerated contours
than some competing methods, it still introduces curved
patterns inconsistent with the HQ image.
In the all-in-one restoration example, the combination of
low light, haze, and rain obscures vegetation and terrain.
\ourmethod{} removes much of the degradation and recovers
clearer scene structure than several competing outputs,
but fine foliage remains smoothed and does not fully
match the reference.

NSD currently constructs its surrogate sample sets within
a single training pair.
A possible improvement is to supplement internal recurrence
with content-matched residuals retrieved from other training
pairs, providing a broader set of restoration examples.
Incorporating uncertainty into sample weighting could
further reduce unreliable detail supervision in ambiguous
regions.
These extensions may improve the balance between detail
recovery and fidelity under challenging degradations.

\end{document}